\documentclass[lettersize,journal]{IEEEtran}

\ifCLASSOPTIONcompsoc
  \usepackage[nocompress]{cite}
\else
  \usepackage{cite}
\fi
\usepackage{bm}
\usepackage{tabularx}
\usepackage{amsthm}
\usepackage{amsmath}
\usepackage{amssymb}
\usepackage{algorithm}
\usepackage{algorithmic}
\usepackage{cite}
\usepackage{multirow}
\usepackage{enumitem}
\usepackage{stfloats}
\usepackage{booktabs,caption}
\usepackage{lipsum}
\usepackage[flushleft]{threeparttable}
\usepackage{graphicx}
\usepackage{xcolor}
\usepackage{mathtools}
\newtheorem{lemma}{Lemma}
\newtheorem{corollary}{Corollary}
\newtheorem{theorem}{Theorem}
\usepackage{booktabs}   % for \toprule, \midrule, \bottomrule
\usepackage{siunitx}
\usepackage{multirow}   % for \multirow
\usepackage{makecell}   % for \makecell (multi-line cells)
\usepackage{array}      % for custom column types
\usepackage[utf8]{inputenc}
\usepackage{amsmath}
\usepackage{amssymb}
\usepackage{graphicx}
\usepackage{booktabs} % For professional looking tables
\usepackage{caption} % For captions
\usepackage{hhline}
\newtheorem{proposition}{Proposition}

\usepackage{comment}
\newcommand{\Esnnc}{E^c_{\text{SNN}}}
\newcommand{\Eannc}{E^c_{\text{ANN}}}
\newcommand{\Nsrc}{N_{\mathrm{src}}}
\newcommand{\Egamma}{\gamma}
\newcommand{\EtotalQNN}{E_{\text{ANN}}}
\newcommand{\EtotalSNN}{E_{\text{SNN}}}

\newcommand{\Emac}{E_{\text{MAC}}}
\newcommand{\Eacc}{E_{\text{ACC}}}
\newcommand{\Ecmp}{E_{\text{CMP}}}
\newcommand{\Esub}{E_{\text{SUB}}}

\newcommand{\Emovedensecustom}{\overline{E}^{\text{move}}}
\newcommand{\Emovesparsecustom}{\widetilde{E}^{\text{move}}}

\usepackage{graphicx}
\usepackage{subcaption}
\usepackage{adjustbox}

\newcommand{\Es}{s_r}
\newcommand{\SR}{s_r}
\newcommand{\Et}{T}

\newcommand{\Ek}{k}
\usepackage{xcolor}
\newcommand{\revise}[1]{{\color{black}#1}}
\newcommand{\revisev}[1]{{\color{black}#1}}

\begin{document}
\title{Reconsidering the Energy Efficiency of Spiking Neural
Networks Inference from Analytical Perspectives}

\author{Zhanglu Yan,
        Zhenyu Bai$^*$, Kaiwen Tang
        and~Weng-Fai Wong,~\IEEEmembership{Senior Member,~IEEE}% <-this % stops a space
\IEEEcompsocitemizethanks{\IEEEcompsocthanksitem Zhanglu Yan, Zhenyu Bai, Kaiwen Tang and Weng-Fai Wong are with the School
of Computing, National University of Singapore. 
\{zhangluyan, wongwf\}@comp.nus.edu.sg; zhenyu.bai@nus.edu.sg

}

%\IEEEcompsocthanksitem J. Doe and J. Doe are with Anonymous University.}% <-this % stops an unwanted space
\thanks{$^*$Corresponding author.}}


\markboth{Journal of \LaTeX\ Class Files}%
{Shell \MakeLowercase{\textit{et al.}}: Bare Demo of IEEEtran.cls for Computer Society Journals}



\IEEEtitleabstractindextext{%
\begin{abstract}
Spiking Neural Networks (SNNs) promise higher energy efficiency over conventional Quantized Artificial Neural Networks (QNNs) due to their event-driven, spike-based computation. However, prevailing energy evaluations often oversimplify, focusing on computational aspects while neglecting critical overheads like comprehensive data movements and memory accesses. 
Such simplifications can lead to misleading conclusions regarding the true energy benefits of SNNs. 
This paper presents a rigorous re-evaluation. We establish a fair baseline by mapping rate-encoded SNNs with $T$ timesteps to capacity-matched QNNs with $\lceil \log_2(T+1) \rceil$ bits. 
This ensures both models have comparable representational capacities, as well as similar hardware requirements, enabling meaningful energy comparisons.
We introduce a detailed analytical energy model encompassing core computation and data movements.  Using this model, we systematically explore a wide parameter space, including intrinsic network characteristics (SNN time window size, spike rate, QNN sparsity, model size, weight bit-level) and hardware characteristics (memory system and network-on-chip). Our analysis identifies specific operational regimes where SNNs genuinely offer superior energy efficiency. For example, under typical neuromorphic hardware conditions, SNNs with moderate time windows ($T = 5$) require an average spike rate ($\SR$) below 5.7\% to outperform equivalent QNNs. %These insights guide the design of energy-efficient neural network solutions. Further, we provide the typical smartwatch lifetime of running SNNs to illustrate the real-world impact.
% Furthermore, to illustrate the real-world implications of our findings, we analyze the operational lifetime of a typical smartwatch, showing that an optimized SNN can nearly double its battery life compared to a QNN. 
These insights guide the design of truly energy-efficient neural network solutions.

\end{abstract}

}

\maketitle

\IEEEdisplaynontitleabstractindextext
\IEEEpeerreviewmaketitle

\section{Introduction}
\label{sec:intro}

\begin{figure}
    \centering
    \includegraphics[width=0.83\linewidth]{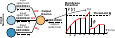}
    \resizebox{0.9\linewidth}{!}{
        \begin{tabular}{c|c|c|c|c}
         T  & 1 & 2 & 3 & 4 \\
         \hline
         Action& $V+w_0+w_2$ & $V+w_0+w_1$ & $V+w_1$ & x   \\
         \hhline{=====}
         T & 5 & 6 & 7 & 8 \\
         \hline
         Action& $V+w_2$; fire; $V-\theta$ & $V+w_0$ & $V+w_1$ & x
        \end{tabular}
    }
    \caption{Integrate-and-fire SNN model}
    \label{fig:if}
\end{figure}

Spiking Neural Networks, characterized by event-driven sparsity and binary spike signaling, represent a compelling approach in neuromorphic computing. Unlike conventional ANNs that process dense tensors, SNNs transmit sparse binary spike trains using {\em integrate-and-fire} (IF) model, significantly reducing computational load and promising substantial energy savings~\cite{ rueckauer2017conversion}. As shown in Figure~\ref{fig:if}, at each timestep \(t\), the SNN neuron calculates the weighted sum of incoming spikes, adds this to a accumulated membrane potential, $V$, from the previous timestep $t-1$, and checks it against a threshold \(\theta\). If the threshold is exceeded, the neuron emits a spike `1' and resets the potential. Otherwise, it outputs a `0'~\cite{stein2005neuronal}. Recent architectures such as Sorbet~\cite{tang2024sorbet} demonstrates successful integration of spiking dynamics in large language models, showcasing notable energy reductions.

However, accurately realizing these energy benefits is complex~\cite{dampfhoffer2022snns}. Prevailing evaluations often rely on simplistic metrics, such as counting the number of additions versus multiplications and assuming a fixed energy consumption per addition and multiplication~\cite{10152465}. The cost of the data movements is one of the critical factors but often overlooked: due to the poor scalability of memories (commonly known as the memory wall~\cite{memory-wall}), the energy consumption of memory operations becomes even dominating over the energy spent by the computations themselves~\cite{horowitz20141}. While SNNs often claim to be event-driven, therefore leveraging spike-train sparsity to skip computations, this mechanism requires multiple data accesses per neuron activation cycles ($T \times \SR$ times per weight, where $T$ is the time window size and $\SR$ is the spike rate), unlike ANNs that typically fetch weights just once per activation. This increased memory access, along with data movements for spikes, can significantly impact the energy consumption, calling the energy efficiency of SNNs into question. Thus, this paper addresses this critical question: \textbf{Under what specific algorithmic and hardware conditions do SNNs achieve genuinely superior end-to-end energy efficiency compared to ANNs?} 

Evaluating the end-to-end energy efficiency of neural network inference is a multifaceted challenge, influenced by interacting factors spanning model design (software), software-to-hardware mapping (or compilation), and the specifics of the underlying hardware architecture and their physical implementations. Ensuring a fair, ``apples-to-apples'' comparison between different neural network paradigms, such as SNNs and conventional ANNs, is therefore inherently complex. To enable a fair comparison, our approach exploits a key observation: each rate-encoding SNN has a near-equivalent Quantized ANN counterpart that we call the \textbf{QNN-SNN twin}. These twins share the same network structure and weight datatype, differing only in their activation representations\revise{, that we demonstrate to have near-equivalent representability}. Leveraging this equivalence enables controlled comparison across varied model settings and hardware settings.

\paragraph{The QNN-SNN Twins}
 Specifically, an SNN operating over $T$ timesteps is paired with a QNN with the same network structure but utilizing activations quantized to $\lceil\log_2(T+1)\rceil$ bits. These QNN-SNN twins share the same network structure and weight datatype, differing only in their activation representation and computation methods. Because of such identicality, we can keep consistent hardware mapping for the twin. Consequently, when comparing twin QNN-SNN pairs, we control both the model capability and mapping, isolating hardware differences and how the twin can be executed on that hardware as the primary factors.

\paragraph{Hardware Energy Consumption}
Hardware inference energy consumption comprises three primary components:
\begin{equation}
E = E_{\text{Compute}} + E_{\text{Data}} (+ E_{\text{Control}})
\label{eq:total_energy}
\end{equation}

\textbf{Compute Energy ($E_{\text{Compute}}$):} This accounts for energy spent on arithmetic operations—additions in SNNs and multiplications plus additions in ANNs—and activation functions. Although SNNs replace multiplications with additions, potentially saving energy due to sparsity, actual energy savings depend on specific hardware implementations, arithmetic designs, operand data types, and the operational sparsity level. 

\textbf{Data Movement Energy ($E_{\text{Data}}$):} Fetching weights and moving activations often use the most energy in inference. The cost depends on two things: the memory and NoC hardware, and how the model is mapped onto it.

% Digital hardware fundamentally relies on moving data among storage (memories) and compute units, performing computations, and subsequently transferring data to the next computation stage or output. Specifically, for SNN and QNN architectures, this includes fetching weights and transferring intermediate activations.

% On digital devices, data movement typically occurs through memory hierarchies or Network-on-Chip (NoC). The resulting data traffic depends heavily on hardware characteristics and the model-to-hardware mapping. Our analysis primarily targets spatial dataflow architectures where data movement predominantly occurs via NoCs rather than memory hierarchies. \footnote{\revise{We consider digital neuromorphic architectures (~\cite{loihi1,akopyan2015truenorth} to be a subset of the widely adopted concept of dataflow architectures (e.g. \cite{tenstorrent,lie2022cerebras,prabhakar2022sambanova}) with specializations.}}

\textbf{Control Energy ($E_{\text{Control}}$):} General-purpose architectures typically invest hardware resources either at the front-end to support unified programming interfaces, such as instruction fetching and decoding for Instruction Set Architectures (ISAs); or at the back-end to enhance performance \revise{sustainability} across diverse applications, \revise{using techniques such as} out-of-order buffers in CPUs and warp schedulers in GPUs. In this study, we exclude these architectural costs related to generality-specialization trade-offs because we mostly focus on specialized architecture where this control cost is low. In other words, we are trying to find out what kind of specialization is more (or less) beneficial for SNN. Instead, \revise{the energy contribution of the control part is ignored in this work. Our study focuses on the energy consumption of the fundamental operations used in SNNs, i.e. the compute and data movement energy, developing a first-principles analytical model rather than a model for any specific architecture. }

\revise{
\paragraph{Objectives}Motivated by the considerations above, the objective of this paper is to develop a first-principles analytical energy model that explicitly accounts for both computation and data movement costs, including weight loading and the loading, storing, and transfer of activations, in order to enable a fair comparison between SNNs and their QNN-SNN twins. Using this model, we analyze a range of hardware and software configurations to identify the spectrum of conditions under which SNNs provide greater energy benefits than their twin QNN counterparts. Ultimately, this work aims to give developers of digital neuromorphic systems a practical way to quickly identify the target operating range in which SNN-based designs are meaningful, enabling fast pre-development validation of whether a given system makes sense from an energy-efficiency perspective.}

\section{\revisev{Representation Capacity-Matched QNN-SNN Twin Construction for Rate-Encoded SNNs}}
\label{sec:equivalence}

\revisev{The purpose of this section is to construct a controlled,
capacity-matched QNN comparator for a rate-encoded SNN. The construction
does not estimate the spike rate, activation sparsity, or energy
consumption of either network. Instead, it fixes the QNN activation
precision required to represent the same $T+1$ output levels as an SNN
operating over $T$ timesteps. Workload-dependent activity statistics,
mapping-dependent action counts, and hardware-dependent energy
parameters are introduced independently in later sections.

We proceed at three levels. First, we establish the construction for one
target IF neuron. Second, we relate the activation density of the
incoming QNN inputs to the input spike rate of the corresponding SNN.
Finally, we aggregate the neuron-level quantities to obtain the
network-level quantities used in the subsequent energy analysis.}

\begin{table}[t]
\centering
\caption{\revise{Symbols used in Theorems and Proofs.}}
\label{tab:symbols_section2}
\footnotesize
\setlength{\tabcolsep}{4pt}
\begin{tabularx}{\columnwidth}{@{}lX@{}}
\hline
\textbf{Symbol} & \textbf{Meaning} \\
\hline
\multicolumn{2}{@{}l}{\textit{Neuron level}} \\
\hline
$v_i^l(t)$ & membrane potential of neuron $i$ at step $t$ \\
$\theta_i^l$ & firing threshold of neuron $i$ \\
$I_i^l(t)$ & net input current of neuron $i$ at step $t$ \\
$s_i^l(t)$ & output spike of neuron $i$ at step $t$ \\
$n_i^l$ & total output spikes of neuron $i$ over $T$ steps \\
$\phi_i^l$ & average firing rate of neuron $i$, $n_i^l/T$ \\
\hline
\multicolumn{2}{@{}l}{\textit{Source-input level}} \\
\hline
$\mathcal{S}_i^{l-1}$ & source-input set of neuron $i$ \\
$N_{\mathrm{src},i}^{l-1}$ & number of source inputs, $|\mathcal{S}_i^{l-1}|$ \\
$k_j^{l-1}$ & total spikes from source input $j$ over $T$ steps \\
$a_j^{l-1}$ & QNN activation of source input $j$ \\
$\gamma_i^{l-1}$ & sparsity over $\mathcal{S}_i^{l-1}$ \\
$s_{r,i}^{\,l-1}$ & average input spike rate into neuron $i$ \\
\hline
\multicolumn{2}{@{}l}{\textit{Network level}} \\
\hline
$s_r$ & network-average spike rate \\
$\gamma$ & network-average activation sparsity \\
 $\mathcal{N}$ & the set of all neurons in the SNN \\
 $N_{\mathrm{src}}$ & network-average number of source inputs\\
\hline
\end{tabularx}
\end{table}

\revise{We first consider a single target neuron and study its \textbf{output representation capability:}}

%In particular, the SNN neuron can realize exactly $T+1$ distinct output levels, and thus its output representation can be matched by a QNN neuron with $\lceil \log_2(T+1)\rceil$-bit activations.

\begin{theorem}
\label{theorem_eq_revised}
%For a rate-encoded IF SNN neuron operating with a time window of size $T$, there exists an equivalent QNN neuron whose activations are quantized to at most $\lceil \log_2(T+1) \rceil$ bits, \revise{such that the two neurons possess comparable information representation capability at the single-neuron output level.}
\revise{Consider a rate-encoded IF SNN neuron operating over a time window of size $T$.
Assume the neuron is non-leaky, uses reset-by-subtraction, and satisfies
$0 \le I_i^l(t) < \theta_i^l$ for every step $t$.
Then there exists a QNN neuron whose output activations are quantized to at most
$\lceil \log_2(T+1)\rceil$ bits such that the QNN exactly matches the single-neuron
output representation of the SNN. \revisev{(Used constructively: it fixes the twin's width
$b=\lceil\log_2(T{+}1)\rceil$; Corollary~1 bounds the non-ideal case.)}

}
\end{theorem}

% We next consider the source-input set of one target neuron, rather than the whole network. 
% Here, $N_{\mathrm{src}}$ denotes the number of incoming inputs connected to that neuron.
\revise{We next consider \textbf{incoming input set of one target neuron}:}

\begin{theorem}
\label{theorem_spikerate_revised}
\revise{Consider the incoming inputs of one target neuron $i$ in the QNN layer $l$, and let $\gamma_i^{l-1}$ denote the sparsity rate over these inputs. }
\revise{Assume that each non-zero QNN activation $a^{l-1}_j \in \{1/T, 2/T,...,1\}$. }
If an SNN is used to represent the same input information with rate coding over a time window $T$, then its average input spike rate $s_{r,i}^{l-1}$ ~\footnote{\revisev{The factor-$T$ width cannot be avoided when only the sparsity $\gamma$ is
known: $\gamma$ fixes how many activations are nonzero, but not how large
they are, and both ends of the interval can really occur. So no tighter
bound is possible from $\gamma$ alone. If the distribution of the nonzero
values is also known, the interval shrinks to a single operating point.}}is bounded by
\[
\frac{1-\gamma_i^{l-1}}{T} \le s_{r,i}^{l-1} \le 1-\gamma_i^{l-1}.
\]

% Consider a target neuron receiving $N_{\mathrm{src}}$ source inputs in the corresponding QNN. Let $\gamma$ denote the sparsity rate over these $N_{\mathrm{src}}$ inputs. Since we establish an equivalence with a rate-coded SNN operating over a time window $T$, we assume that each non-zero QNN input activation $a_j$ lies in the range $[1/T, 1]$. Here, $a_j=1/T$ signifies the minimal non-zero signal (equivalent to one spike event in the SNN over $T$ timesteps), and $a_j=1$ represents the maximal signal (equivalent to SNN events at every discrete opportunity within $T$). If an SNN is to represent the same input information as this QNN using rate coding, its average input spike rate $\SR$ (defined as the average number of spikes per input neuron per SNN timestep) is bounded by:
% \[
% \frac{1 - \gamma}{T} \leq \SR \leq 1 - \gamma.
% \]
% %This bound characterizes the relationship between the QNN's input activation density and the corresponding SNN's average input spike rate.
\end{theorem}

\revise{Theorem~1 and Theorem~2 are stated for one target neuron and its incoming input set. 
We now extend these local results to the whole network. 
Since the same local construction applies independently to every neuron under the same network architecture and time window $T$, the network-level result follows by applying the neuron-wise construction to all neurons and aggregating the local quantities.}

\begin{proposition}[Network-level extension]
\label{prop:network_extension}
Let $\mathcal{N}$ denote the set of all neurons in the SNN.

\revise{
\textbf{(a) Extension of Theorem~1.}
If Theorem~1 holds for every neuron $(i,l)\in\mathcal{N}$, then there
exists a QNN twin with the same network architecture such that each
QNN neuron uses at most $\lceil \log_2(T+1)\rceil$-bit activations and
matches the output representation of its corresponding SNN neuron. 

\textbf{(b) Extension of Theorem~2.}
For every neuron $(i,l)\in\mathcal{N}$, Theorem~2 gives
$\frac{1-\gamma_i^{l-1}}{T}
\le
s_{r,i}^{\,l-1}
\le
1-\gamma_i^{l-1}.
$
Aggregating these neuron-level quantities over the whole network with
weights $N_{\mathrm{src},i}^{l-1}$ yields
\[
s_r=
\frac{\sum\nolimits_{(i,l)\in\mathcal N} N_{\mathrm{src},i}^{\,l-1} s_{r,i}^{\,l-1}}
{\sum\nolimits_{(i,l)\in\mathcal N} N_{\mathrm{src},i}^{\,l-1}},
\!\gamma=
1-\frac{\sum\nolimits_{(i,l)\in\mathcal N} N_{\mathrm{src},i}^{\,l-1}(1-\gamma_i^{\,l-1})}
{\sum\nolimits_{(i,l)\in\mathcal N} N_{\mathrm{src},i}^{\,l-1}}.
\]
and therefore
$
\frac{1-\gamma}{T}
\le
s_r
\le
1-\gamma.
$}
%That is, the sparsity--spike-rate relation in Theorem~2 is preserved from the neuron level to the whole network under% averaging.
\end{proposition}
% \begin{proposition}
% \label{prop:network_extension}
% \revise{Let $\mathcal{N}$ denote the set of all neurons in the SNN.
% \textbf{(a) Extension of Theorem~1.}
% If Theorem~1 holds for every neuron $(i,l)\in\mathcal{N}$, then there exists a QNN twin with the same network architecture such that each QNN uses at most $\lceil \log_2(T+1)\rceil$ bits and matches the output representation of its corresponding SNN. \textbf{(b) Extension of Theorem~2.}
% For each neuron $(i,l)\in\mathcal{N}$, let $\gamma_i^l$ and $s_{r,i}^{\,l-1}$ denote the sparsity rate and average input spike rate over its incoming input set, and let $N_{\mathrm{src},i}^{l-1}$ denote the number of incoming inputs. Then for every neuron $(i,l)\in\mathcal{N}$. Multiplying by $N_{\mathrm{src},i}^{l-1}$, summing over all neurons, and normalizing by $\sum_{(i,l)\in\mathcal{N}} N_{\mathrm{src},i}^{l-1}$ yields
% \[
% \frac{1-\gamma}{T}
% \le
% s_r
% \le
% 1-\gamma.
% \]
% }
% \end{proposition}
 \revise{The details of Proof for Theorem~1, Theorem~2 and Proposition~1 are shown in Appendix~\ref{sec: proof} to~\ref{proofC}.}

\section{Hardware}
\label{sec:hardware}
Section~\ref{sec:equivalence} established the relationship between a twin of SNN and QNN which have near-equivalent representational capabilities. 
 This equivalence maintains consistency in weight datatypes and overall network architecture, isolating differences to the representation of activations and the computations required to generate them. Building on this foundation, this section delves into the \revise{energy model of the QNN-SNN twin with} hardware implications arising from these distinctions. Our primary objective is to deduce the energy trade-offs between SNNs and QNNs, thereby identifying the specific operational conditions and parameter regimes under which one paradigm achieves superior energy efficiency over the other. \revise{To simplify the explanation, i}n the following sections, we consider the energy consumption of a single output neuron with $N_{\text{src}}$ input neurons for both the SNN and the QNN forming a twin SNN-QNN pair, \revise{knowing that the energy consumption of a layer can be calculated with its layer width and its previous layer's width; and the end-to-end energy consumption is calculated by the sum of the each layer}. Specifically, the SNN with $T$ timestep is paired to a QNN with $\lceil \log_2(T+1)\rceil$ bits-integer quantization. The latter processes its input in one forward pass with integer multiplication and accumulation while the former spreads computations over $T$ time steps with additions only. Although SNN are known to be sparse (or the \textit{spike rate} as the density), QNN may also have sparsity due to the sparsification effect of the activation function such as $ReLU$. Let $\gamma$ be the fraction of zero activations in the QNN (sparsity), and $\SR$ the average spike rate (density) per input in the SNN.  %Our analytical framework is indeed network-agnostic. The structural differences between CNNs and Transformers are captured by our model's core parameters, primarily $N_{src}$ (e.g., input channel numbers$ \times$ kernel width and length, for a CNN vs. the FFN dimension for a Transformer).

\subsection{Core computing $(E_{Compute})$}
\label{sec: compute}
Core computing energy quantifies the energy consumed by fundamental arithmetic operations. For Artificial Neural Networks, this primarily involves multiply-accumulate (MAC) operations, whereas Spiking Neural Networks predominantly utilize accumulate (ACC) operations. 

In a typical ANN layer, core computation consists of summing $\Nsrc$ weighted inputs, followed by a non-linear activation function. This activation often includes clamping to constrain output values to a defined range(e.g., [0,1]). The computational energy for ANNs, $\Eannc$, can thus be expressed as:
\begin{equation}
\label{eq:ann_compute_energy}
\Eannc = \Nsrc \cdot \underbrace{(1-\Egamma) \cdot \Emac}_{\text{Active Operations}} + \underbrace{2\Ecmp}_{\text{Clamping}}
\end{equation}
where $\Emac$ is the energy per MAC operation, and $\Ecmp$ is the energy per comparison and clamping. Due to the sparsity, only $(1-\Egamma)$ computations are necessary after skipping the zero activations.
% For SNNs employing Integrate-and-Fire neuron models, MACs are replaced by additions (accumulations). The core computational energy for SNNs, $\Esnnc$, is given by:
% \begin{multline}
% \label{eq:snn_compute_energy}
% \Esnnc = \underbrace{\Nsrc \cdot \Et \cdot \SR \cdot \Eacc}_{\text{Membrane Potential Update}} + \underbrace{\Et \cdot (\Ecmp + \SR \cdot \Esub)}_{\text{Spiking Operations over T}}
% \end{multline}
% Here, $\Et$ is the number of timesteps, $\Es$ is the average spike rate, $\Eacc$ is the energy for an accumulation, and $\Esub$ is the energy for a subtraction\revise{, representing the} membrane potential update after firing.
For SNNs employing Integrate-and-Fire neuron models, MACs are replaced by additions (accumulations). \revisev{We additionally account for one local membrane-state read and one local membrane-state write per output neuron at every timestep.} The core computational energy for SNNs, $\Esnnc$, is given by:
\begin{multline}
\label{eq:snn_compute_energy}
\Esnnc =
\underbrace{\Nsrc \Et \SR \Eacc}_{\text{Membrane Update}}
+ \underbrace{\Et(\Ecmp+\SR\Esub+\revisev{E^V_{\mathrm{rd}}+E^V_{\mathrm{wr}})}}_{\text{Spiking Operations over T}}
.
\end{multline}
Here, $\Et$ is the number of timesteps, $\Es$ is the average spike rate, $\Eacc$ is the energy for an accumulation, and $\Esub$ is the energy for a subtraction\revise{, representing the} membrane potential update after firing. 

To compare the computational energy efficiency, we introduce a parameter $\Ek$, representing the energy ratio of a MAC operation over an addition. Assuming $E_{\text{ACC}}=E_{\text{CMP}}=E_{\text{SUB}}$, we define $k=\Emac/E_{\text{ACC}}$. An SNN offers computational energy savings if $\Esnnc\leq\Eannc$, which \revisev{now leads to the condition}
\begin{equation}
\label{eq:snn_adv_condition}
\Et\Es+
\frac{\Et+\Et\Es-2+\revisev{\Et(E^V_{\mathrm{rd}}+E^V_{\mathrm{wr}})/\Eacc}}
{\Nsrc}
\leq \Ek(1-\Egamma).
\end{equation}
\revisev{The second term is a per-output-neuron overhead and is progressively amortized as $\Nsrc$ grows. Dropping this $O(1/\Nsrc)$ term recovers the high-level intuition $\Et\SR\lesssim\Ek(1-\Egamma)$. This simplification is used only to discuss qualitative best-, average-, and worst-case trends; all numerical results use the full expressions in Eqs.~\eqref{eq:ann_compute_energy} and \eqref{eq:snn_compute_energy}, including membrane-state accesses.} We analyze this simplified condition under three SNN spike rate scenarios discussed in Theorem~\ref{theorem_spikerate_revised}:

% To compare the computational energy efficiency, we introduce a parameter $\Ek$, representing the energy ratio of a MAC operation over an addition. Assuming $E_{\text{ACC}} = E_{\text{CMP}} = E_{\text{SUB}}$, we define $k= \Emac/E_{\text{ACC}}$. An SNN offers computational energy savings if $\Esnnc \leq \Eannc$, which leads to the condition:
% \begin{equation}
% \label{eq:snn_adv_condition}
% \Et\Es + \frac{\Et + \Et\Es - 2}{\Nsrc} \leq \Ek(1 - \Egamma)
% \end{equation}

% For typical large-scale neural networks, $\Nsrc$ \revise{(which is equal to the pevious layer's width)} is large (e.g., $\ge 10^3$), rendering the term $\frac{\Et + \Et\Es - 2}{\Nsrc}$ negligible. This simplifies the condition for SNN computational energy advantage to $\Et\Es \leq \Ek(1 - \Egamma)$. \footnote{This simplification was used only in Section III.A to provide a high-level intuition; in the later results section, all concrete energy calculations for both ANNs and SNNs were computed using the full, unsimplified energy models presented in Eq.~(2) and Eq.~(3).} We analyze this simplified condition under three SNN spike rate scenarios discussed in Theorem~\ref{theorem_spikerate_revised}:

\begin{itemize}
    \item \textbf{Best-case for SNN ($\Es = \frac{1 - \Egamma}{\Et}$):} Corresponds to the minimum SNN spike rate to match QNN activation density $(1-\Egamma)$ over $\Et$ timesteps. SNNs are more computationally energy-efficient if $\Ek \geq 1$ \revise{,i.e. as soon as multiplication consumes extra energy, which is generally true}.
    \item \textbf{Average-case for SNN ($\Es = \frac{(1 - \Egamma)(1/\Et + 1)}{2}$):} SNNs are favored if $\Ek \geq \frac{1 + \Et}{2}$. \revise{That says, the SNNs are energy saving if the energy of multiplication is greater than $\frac{1 + \Et}{2} - 1 = \frac{\Et -1}{2}$ of the energy of an addition.}
    \item \textbf{Worst-case for SNN ($\Es = 1 - \Egamma$):} Represents the maximum SNN spike rate, where each potential input effectively triggers a spike in every relevant timestep. SNNs are favored if $\Ek \geq \Et$. (The referenced study notes $\Ek>\Et$ for strict superiority). \revise{That is, if SNNs have too many spikes, they can win only if the multiplication takes a lot of energy and the number of time steps is small. }
\end{itemize}

%This structured analysis provides clear thresholds and guidance for determining the conditions under which the use of SNN architectures yields tangible energy efficiency advantages over conventional ANN implementations.

\subsection{Data Movement Energy ($E_{data}$)}
\label{sec:data_movement_energy}

\revisev{As motivated in Section I, data movement—fetching weights and transferring activations—often dominates inference energy; its cost depends on the memory/NoC technology and the model-to-hardware mapping.} \revise{Many digital neuromorphic architectures, such as Loihi~\cite{lines2018loihi} and TrueNorth~\cite{akopyan2015truenorth}, can be viewed as specialized forms of dataflow systems. Although they are often designed primarily for spiking neural networks, they share important architectural principles with other dataflow systems which target a broader range of AI and high-performance computing workloads. The goal of our energy model is not to construct highly detailed, architecture-specific models, but rather to establish a first-principles understanding of the fundamental operations involved in data movement. Accordingly, we focus on the general energy costs of data movement in dataflow architectures, using this term hereafter to include digital neuromorphic architectures as well.}

%Their very distinction of neuromorphic architectures often lies in more specialized control/compute components tailored for SNN operations. 

\begin{figure}[htbp]
    \centering
    % Ensure the path to your figure is correct
    \includegraphics[width=0.6\linewidth]{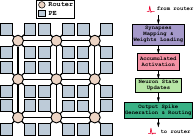} 
    \caption{A classical neuromorphic processing element (PE) array with a Network-on-Chip (NoC) for inter-core communication~\cite{lines2018loihi,northpole,pehle2022brainscales}.}
    \label{fig:neuromorphic-arch}
\end{figure}

% Since dataflow architectures generally require static mapping, we consider a common mapping strategy for SNN where weights are stored locally within each \textit{core's} SRAM, and activations are shared across cores via a dedicated NoC, as illustrated in the archetypal architecture shown in Figure~\ref{fig:neuromorphic-arch}. Each core integrates local memory, control logic, and compute units for ordered task execution. This is also the reason we priotize dataflow architecture over other spatial architectures such as GPUs because the latters have complex dynamic schedule and memory behavior that are hard to be controlled for a fair and accurate comparison between the SNN-QNN twins.
% Although this \textit{weight-stationary} mapping is not necessarily the optimal mapping for SNN or QNN depending on the datatype and the model architecture, this mapping 
% can ensure a fair comparison between SNN and the twin QNN.  
\revise{Since our target platforms are dataflow-style architectures, we assume a static mapping of the network onto hardware resources, as is common in prior SNN compiler and mapping flows for predictable execution and communication analysis \cite{akopyan2015truenorth,loihi1}. As illustrated in Figure~\ref{fig:neuromorphic-arch}, each core contains local memory, control logic, and compute units, while inter-core communication is handled through an on-chip network. Under this mapping, synaptic weights and neuron state are kept in each core's local SRAM, and spikes/activations are transmitted across cores via the NoC. This organization is consistent with representative neuromorphic processors such as IBM TrueNorth, whose neurosynaptic cores communicate through an event-routing fabric, and Intel Loihi, whose manycore design maintains local state within cores and exchanges spike messages over an on-chip interconnect \cite{akopyan2015truenorth,loihi1}. }
Accordingly, we adopt this weight-stationary, local-state mapping as the common
baseline for both the SNN and its twin QNN: while not necessarily
optimal for every architecture or datatype, it fixes data placement
and inter-core communication identically across the pair, enabling
a directly comparable evaluation.

%Accordingly, we adopt this weight-stationary, local-state mapping as the common baseline for both the SNN and its twin QNN. Although this mapping is not necessarily optimal for every model architecture or datatype, it makes data placement and inter-core communication explicit and directly comparable across the two twins. This follows the same principle used in spatial accelerator studies, where different dataflows are compared under identical hardware constraints to enable fair evaluation. By contrast, GPU execution depends much more strongly on dynamic thread scheduling and cache behavior, which are less explicit and therefore harder to control in a tightly matched SNN--QNN comparison \cite{ibrahim2020analyzing}.

\revise{We need to further distinguish the energy consumption of moving activations by separating sparse and dense activation movements. We denote $\Emovedensecustom$ and $\Emovesparsecustom$ as the energy cost per bit moved for dense and sparse data movements, respectively, where typically $\Emovesparsecustom > \Emovedensecustom$. The reason is that sparse activations are often transported as fine-grained, dynamically generated events over the on-chip interconnect, so the useful payload is accompanied by additional transaction overheads such as packet headers, routing and arbitration, buffer accesses, and control activity. When the same amount of useful information is fragmented into many small transfers, these per-transaction costs are amortized less effectively, leading to a higher energy cost per useful bit moved~\cite{dally2004principles, benini2002networks,jerger2009chip}. This assumption is particularly relevant for neuromorphic architectures in which spike/event communication is carried by an on-chip network rather than by a purely streaming dense datapath~\cite{akopyan2015truenorth,loihi1}. For weights, the energy per access ($E^{\text{weight}}$) is assumed identical for SNNs and their corresponding QNNs due to matched network architecture and weight precision; the primary difference lies in the \textit{number} of weight accesses.}

%We need to further distinguish the energy consumption of moving activations by moving sparse activation and dense activation. We denote $\Emovedensecustom$ and sparse $\Emovesparsecustom$ for the energy cost per bit moved, respectively for dense data movements and sparse data movements, where typically $\Emovesparsecustom > \Emovedensecustom$ due to the overheads of finer-grained dynamic data handling. For weights, the energy per access ($E^{\text{weight}}$) is assumed identical for SNNs and their corresponding QNNs due to matched network architecture and weight precision; the primary difference lies in the \textit{number} of weight accesses.

The data movement energy for QNNs, considering both sparse ($\widetilde{E}^d_{\text{QNN}}$) and dense ($\overline{E}^d_{\text{QNN}}$) activation transfers, is formulated as:
\begin{align}
    \widetilde{E}^d_{\text{QNN}} &= \Nsrc \cdot (1 - \Egamma) \cdot (\lceil\log_2 (\Et+1)\rceil \cdot \Emovesparsecustom + E^{\text{weight}}) \label{eq:ann_data_sparse} \\
    \overline{E}^d_{\text{QNN}} &= \Nsrc \cdot (\lceil\log_2 (\Et+1)\rceil \cdot \Emovedensecustom + E^{\text{weight}}) \label{eq:ann_data_dense}
\end{align}
Similarly, for SNNs, which transmit 1-bit spikes over $\Et$ timesteps with an average spike rate $\Es$~\footnote{\revise{Average spike rate is a first-order proxy for event volume; in more general SNN systems, burstiness, spatial imbalance, and memory/NoC effects can further affect communication cost. Under the bounded IF regime assumed in Theorem~1, however, spike generation is well controlled, making average spike rate a reasonable first-order metric in our analysis.}}:
\begin{align}
    \widetilde{E}^d_{\text{SNN}} &= \Nsrc \cdot \Et \cdot \Es \cdot (\Emovesparsecustom + E^{\text{weight}}) \label{eq:snn_data_sparse} \\
    \overline{E}^d_{\text{SNN}} &= \Nsrc \cdot \Et \cdot (\Emovedensecustom + E^{\text{weight}}) \label{eq:snn_data_dense}
\end{align}
We consider the actual data movement energy for each network, $E^d_{\text{QNN}}$ and $E^d_{\text{SNN}}$, is the minimum of its sparse and dense options, i.e. using sparsity only if it is beneficial.

Comparing dense transfers (Eq.~\eqref{eq:ann_data_dense} and Eq.~\eqref{eq:snn_data_dense}), $\overline{E}^d_{\text{QNN}}$ is generally lower than $\overline{E}^d_{\text{SNN}}$ due to the QNN processing $\lceil\log_2 (\Et+1)\rceil$-bit activations once, while the SNN effectively processes $\Et$ individual 1-bit potential spike slots. Thus, SNN energy advantages in data movement hinge on effective sparsity utilization, as captured by $\widetilde{E}^d_{\text{SNN}}$ (Eq.~\eqref{eq:snn_data_sparse}).
To analyze when SNNs are more energy-efficient than QNNs in data movement, we compare the sparse SNN transfer in Eq.~\eqref{eq:snn_data_sparse} with the sparse and dense QNN transfers in Eqs.~\eqref{eq:ann_data_sparse} and \eqref{eq:ann_data_dense}. 
In the following section, we analyze SNNs under the best, average, and worst cases defined in Section~\ref{sec: compute}.

\section{Results}
\label{sec:results}
This section presents the quantitative analyses of energy consumption based on the analytical formulas of each part from the previous section. We will use the ratio between the SNN energy and its twin QNN energy $\EtotalSNN/\EtotalQNN$ as the primary factors to indicate the SNN's energy efficiency.

\subsection{Energy Efficiency Landscape Across Diverse Configurations}
\label{sec:landscape_interpretation}
In this section, we evaluate the combinations of three typical software scenarios and three hardware settings; then compare the energy efficiency of SNNs and QNNs under these conditions.
The three SNN model configurations vary in their time window size $T$ and spike rate $\SR$ while the size of models are fixed to $N_{src}=4096$ ~\cite{touvron2023llama}. We note that with respect to the changes in $T$ and $\SR$, the bit-width of the QNN changes accordingly (Theorem~\ref{theorem_eq_revised} and Theorem~\ref{theorem_spikerate_revised}, respectively).
These cases are chosen to model scenarios ranging from highly efficient, sparse SNNs to less optimized, more active networks, with the typical case reflecting parameters common in state-of-the-art literature~\cite{9556508,10843312,zhou2023spikingformer,han2020deep,han2020rmp}. In practice, these different model parameters need to be chosen with respect to the deployment power constraint and the application requirement of the model performance. Concretely, the three cases are:

\begin{enumerate}
    \item \textbf{Efficient SNN ($T=2, \SR=0.02$):} Models an idealized SNN with a short time window and very low spike rate, representing a best-case scenario for SNN energy performance.
    \item \textbf{Typical SNN ($T=4, \SR=0.1$):} Reflects parameters common in well-known SNN architectures like DIET-SNN~\cite{9556508}, VR-SNN~\cite{10843312}, and Spikingformer~\cite{zhou2023spikingformer}, serving as a realistic baseline.
    \item \textbf{High-Performance SNN ($T=32, \SR=0.20$):} This configuration represents SNNs where achieving high task accuracy is prioritized, a goal that often correlates with longer integration windows and higher spike rates. For instance, state-of-the-art models like Sorbet~\cite{tang2024sorbet} achieve high performance on GLUE benchmarks while operating with a spike rate of approximately 20\% ($\SR=0.2$). This scenario allows us to analyze the energy implications of such accuracy-focused SNN designs. For these CV tasks, TCS-SNN and RMP-SNN both adapt time window size of 32-64 to achieve a higher detecting accuracy~\cite{han2020deep,han2020rmp}. 
\end{enumerate}

%Further, in Figure~\ref{fig:main_summary_energy_landscape_revised}, we compare the SNN against three QNN counterparts, corresponding to ``Worst,'' ``Average,'' and ``Best'' cases for the SNN's relative efficiency as discussed in Theorem~\ref{theorem_spikerate_revised}. These cases are defined by the QNN's activation sparsity ($\gamma$), which is derived from the SNN's $\SR$ and $T$ as $\gamma=1-\SR$ (Worst), $\gamma =1- \frac{2\SR T}{T+1}$ (Average), and $\gamma=1-\SR T$ (Best), respectively.

%Figure~\ref{fig:main_summary_energy_landscape_revised} shows the total of nine possible combinations. Within each of the nine subplots, we compare the SNN against three QNN counterparts, corresponding to ``Worst,'' ``Average,'' and ``Best'' cases for the SNN's relative efficiency as discussed in Theorem~\ref{theorem_spikerate_revised}. These cases are defined by the QNN's activation sparsity ($\gamma$), which is derived from the SNN's $\SR$ and $T$ as $\gamma=1-\SR$ (Worst), $\gamma =1- \frac{2\SR T}{T+1}$ (Average), and $\gamma=1-\SR T$ (Best), respectively. 

We also consider three hardware configurations: a theoretical minimum (to isolate algorithmic costs), a typical neuromorphic setting (to model a realistic implementation), and a theoretical worst sparse processing case (a condition least favorable to SNNs). 
\begin{enumerate}
    \item \textbf{Theoretical Minimum:} Sets all data movement and dynamic control costs to zero ($\overline{E}^{\text{move}}=0, \widetilde{E}^{\text{move}}=0$), isolating the fundamental energy of computation and static weight access.
    \item \textbf{Typical Neuromorphic:} This setting models a realistic deployment on specialized hardware. We adopt a sparse data movement energy cost inspired by Intel's Loihi neuromorphic chip, setting $\widetilde{E}^{\text{move}}=3.0 \text{ pJ/bit/hop}$. This number is an averaged energy cost reported for Loihi 1~\cite{loihi1}, including the overhead for transferring metadata such as the routing information. The dense data movement cost, $\overline{E}^{\text{move}}=0.25 \text{ pJ/bit/hop}$, is based on our own measurements for a circuit-switch NoC on a commercialized 22nm process. We assume a mapping where adjacent layers are placed on adjacent cores, resulting in a hop count of 1.

    \item \textbf{Worst-Case for Sparse Processing:} This scenario models a hardware configuration fundamentally hostile to event-driven processing, representing the least favorable condition for SNNs. To model this high-energy context, we assume that every 1-bit spike necessitates a full 64-bit word read from off-chip DRAM, the most energy-intensive memory tier in our analysis. Based on energy values for a 64-bit DRAM access (approx. 1300 pJ~\cite{horowitz20141}), the effective per-bit cost for a sparse access becomes an extreme $\widetilde{E}^{\text{move}}=1300 \text{ pJ/bit}$. The corresponding dense data movement cost is the amortized value, $\overline{E}^{\text{move}}=1300/64 \approx 20.3 \text{ pJ/bit}$.
\end{enumerate}

\begin{figure}[]
    \centering
    \includegraphics[width=\linewidth]{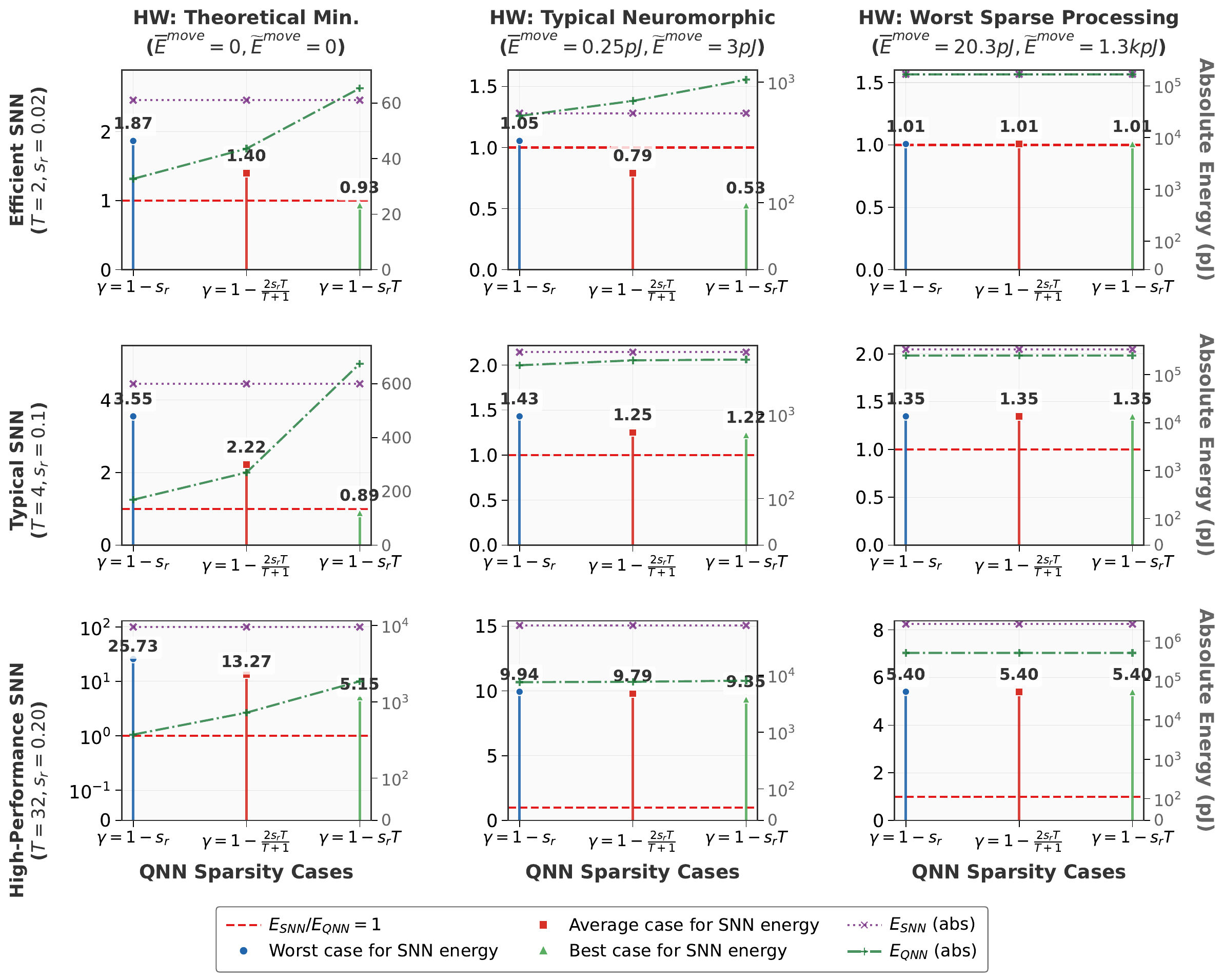} % Replace with your actual figure
    \caption{Energy ratio $\EtotalSNN/\EtotalQNN$ across SNN Model Configurations (rows, defined by $T, \SR$) and Hardware Settings (columns). Within each cell, three bars correspond to comparing the SNN against QNNs with three activation densities. All calculations assume $N_{\text{src}}=4096$ and 8-bit weights. Fundamental operational costs are: $\Eacc=0.05448 \text{ pJ}$, $\Ecmp=0.05448 \text{ pJ}$, $\Esub=0.05448 \text{ pJ}$. \revisev{we use one value for all three
    operations for simplicity, which slightly favors the QNN.} The QNN $\Emac$ cost varies with $T$ as detailed in Figure~\ref{fig:mac_acc}. These energy values are based on a 22nm technology process.}
    \label{fig:main_summary_energy_landscape_revised}
\end{figure}

\begin{figure}[]
    \centering
    \includegraphics[width=.8\linewidth]{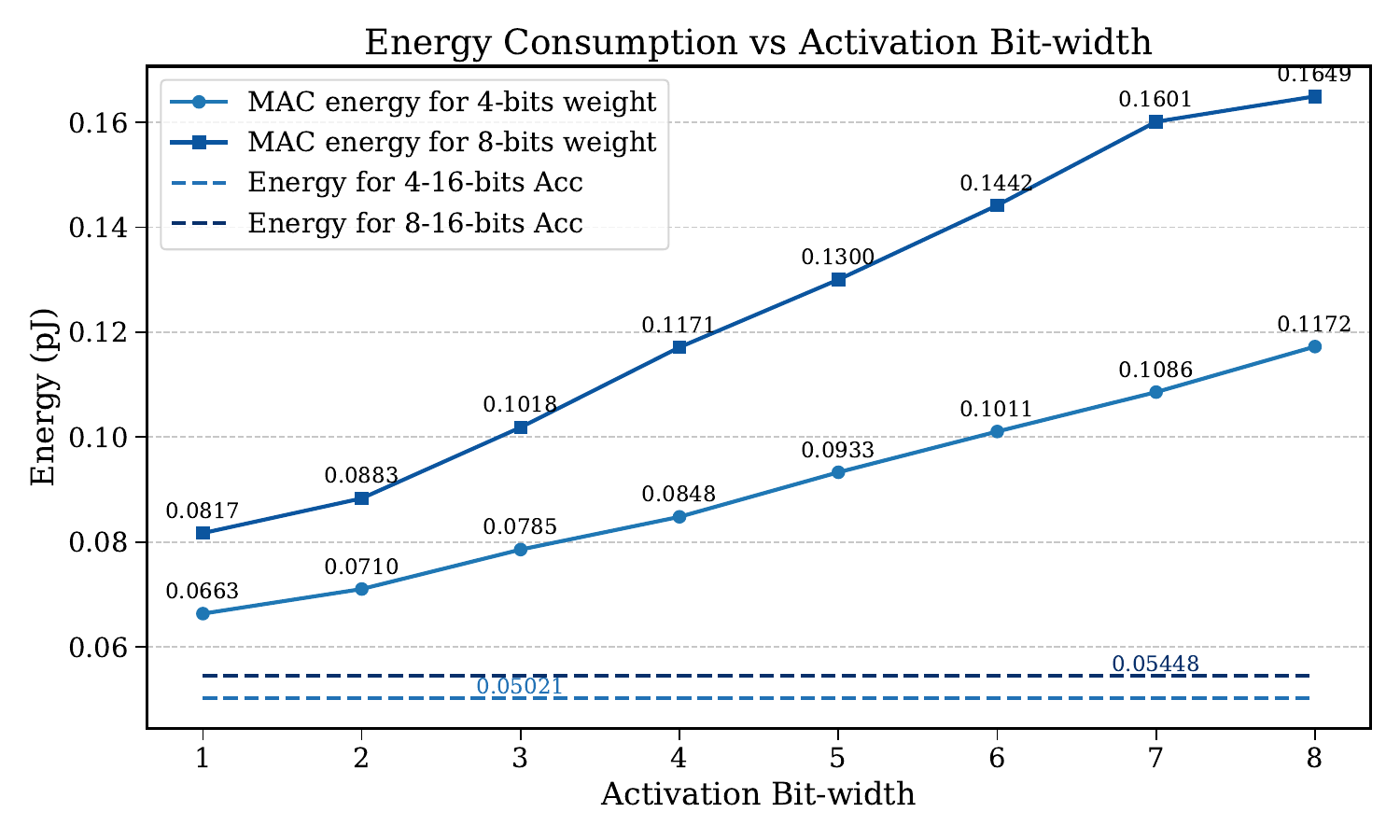}
    \caption{Per-operation energy
of MAC vs.\ ACC as a function of activation bit-width}
    \label{fig:mac_acc}
\end{figure}

For efficient SNN configurations (top two rows), the energy advantage over a QNN (the ``Best case'' scenario) can actually increase when moving from the Theoretical Minimum to the Typical Neuromorphic setting. This occurs because the added data movement costs penalize the multi-bit QNN activations compared to the 1-bit SNN spikes. However, any potential SNN advantage is eliminated and reversed under the Worst-sparse-processing hardware scenario, where extreme overheads on sparse processing overwhelm SNN benefits.  This landscape view clearly illustrates that SNN energy efficiency is not absolute but a conditional property emerging from a complex interplay of algorithmic choices and hardware realities, motivating the fine-grained parameter study in the next section.

\subsection{Sensitivity Analysis of Neural Network Parameters} % Or \subsection if part of a larger Results section
\label{sec:sensitivity_analysis}

\begin{figure}[htbp!]
    \centering
    \includegraphics[width=1.03\linewidth]{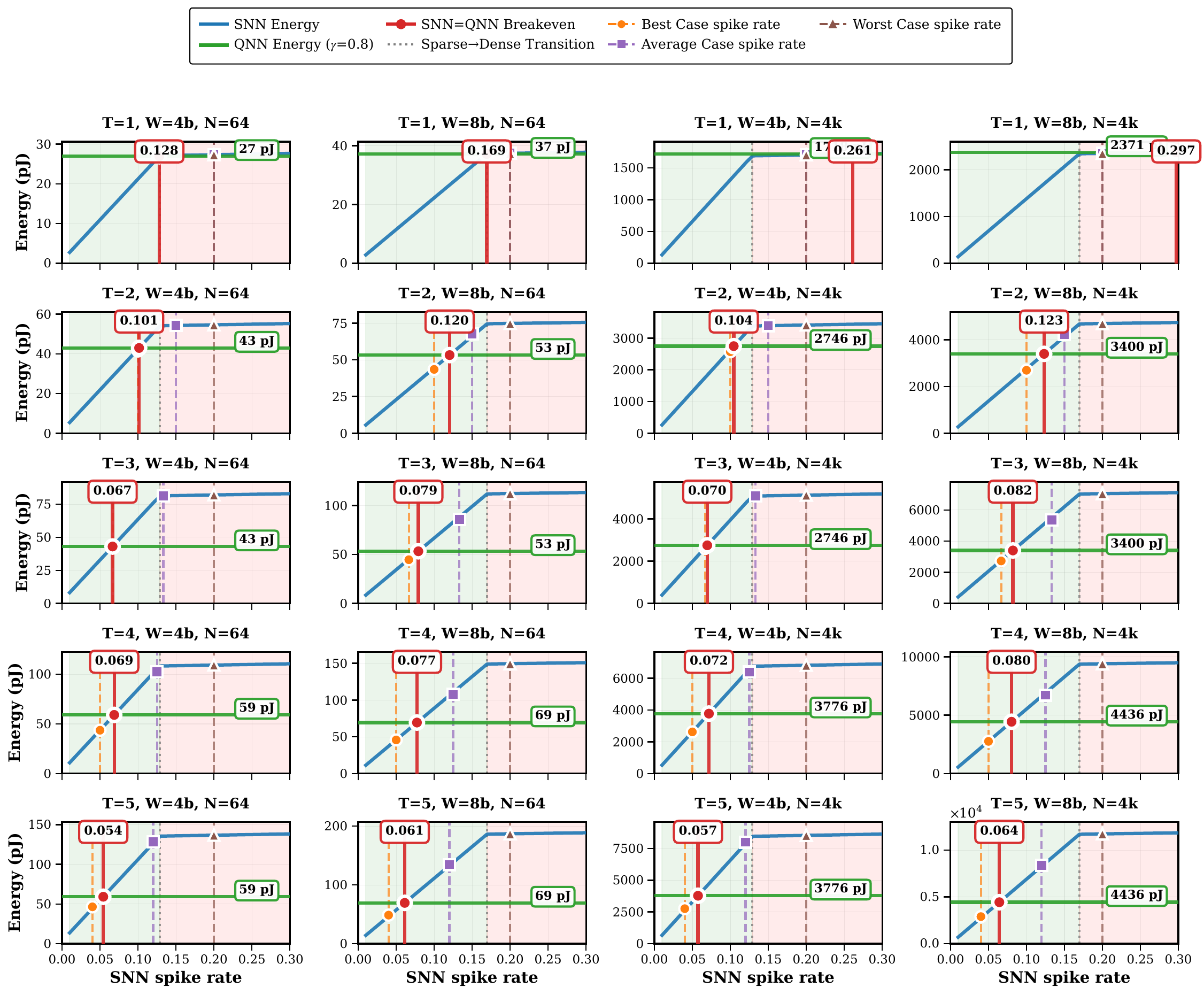} % Replace with your actual figure
    \caption{ Sensitivity analysis of SNN and QNN energy consumption (pJ) versus SNN spike rate ($\SR$) under the Typical Neuromorphic hardware setting ($\overline{E}^{\text{move}}=0.25 \text{ pJ/bit/hop}$, $\widetilde{E}^{\text{move}}=3 \text{ pJ/bit/hop}$, number of hop equal to 1). Each subplot corresponds to a specific SNN time window $T$ (rows, $T \in [1,5]$) and a configuration of weight precision (4-bit or 8-bit) and network size ($N_{\text{src}} \in \{64, 4096\}$) (columns). QNN energy (horizontal line) is calculated for a fixed activation sparsity of $\gamma=0.8$.
}
    \label{fig:sensitivity_detail_plots}
\end{figure}

Following the broad overview of the energy landscape, this section presents a fine-grained sensitivity analysis to quantify the impact of individual neural network parameters on energy consumption. For this study, we fix the hardware assumptions to the Typical Neuromorphic setting, as it represents a realistic deployment scenario for specialized accelerators. This allows us to isolate the effects of intrinsic model characteristics. Against a fixed, optimized QNN baseline with 80\% activation sparsity~\cite{gong2024fast}, we systematically vary key SNN parameters—time window ($T$), spike rate ($s_r$), weight precision (4-bit and 8-bit), and network size ($N_{\text{src}}$)—to quantify their influence on the energy ratio $E_{\text{SNN}}/E_{\text{QNN}}$. Figure~\ref{fig:sensitivity_detail_plots} presents the results of this parametric sweep.

% \textbf{Impact of Time Window ($T$) and SNN Spike Rate ($\SR$):} A primary observation is that as $T$ increases, the SNN must operate at an increasingly lower spike rate ($\SR$) to remain energy-competitive with the QNN. For instance, with 8-bit weights and $N_{\text{src}}=4096$, the breakeven point where $E_{\text{SNN}} = E_{\text{QNN}}$ occurs at $\SR \approx 0.3$ for $T=1$, but this threshold tightens significantly to $\SR \approx 0.082$ for $T=3$. This trend is expected, as the SNN's dominant energy costs scale with the $T \cdot \SR$ product, while the QNN's energy cost increases more slowly with the QNN precision of $\lceil \log_2(T+1) \rceil$.

\revisev{\textbf{Impact of Time Window ($T$) and SNN Spike Rate ($\SR$):} As $T$ increases, the SNN must operate at an increasingly lower spike rate to remain energy-competitive with its QNN twin. With 8-bit weights and $N_{\text{src}}=4096$, the breakeven point is $\SR\approx0.297$ at $T=1$ and decreases to $\SR\approx0.082$ at $T=3$. The dominant event-dependent SNN terms scale with $T\SR$, whereas the QNN activation precision grows only as $\lceil\log_2(T+1)\rceil$; the membrane-state term adds a further cost per timestep.}

\revisev{The only local non-monotonicity occurs between $T=3$ and $T=4$ for 4-bit weights and $N_{\text{src}}=64$: the breakeven spike rate changes from $\SR\approx0.067$ to $\SR\approx0.069$. This small relaxation occurs because the capacity-matched QNN precision increases from 2 to 3 bits when $T$ changes from 3 to 4, producing a step increase in $E_{\text{MAC}}$. The overall trend remains unchanged: increasing $T$ generally tightens the spike-rate requirement for SNN energy savings.}

Furthermore, Figure~\ref{fig:sensitivity_detail_plots} highlights a critical threshold for the SNN's internal data transmission strategy. For the Typical Neuromorphic setting, this threshold occurs at $\SR \approx 0.12-0.17$, depending on weight bit-width. When the SNN spike rate $\SR$ exceeds this value, the energy cost of handling sparse events ($N_{\text{src}} T \SR (\widetilde{E}^{\text{move}} + E_{\text{weight}})$) surpasses that of broadcasting data densely ($N_{\text{src}} T (\overline{E}^{\text{move}} + E_{\text{weight}})$). This indicates that for higher-activity SNNs, a dense dataflow can be more energy-efficient than a sparse, event-driven one within the same architecture.

\revisev{\textbf{Impact of Network Size ($N_{\text{src}}$):} Absolute synaptic energy scales with $N_{\text{src}}$, whereas the membrane-state read/write term is paid once per output neuron and timestep. Consequently, the state cost is better amortized at large fan-in. For example, at $T=5$ with 4-bit weights, the breakeven spike rate is $\SR\approx0.057$ for $N_{\text{src}}=4096$ and $\SR\approx0.054$ for $N_{\text{src}}=64$. Thus, network size has only a modest effect in the large-fan-in regime used for our headline result, but the fixed state cost becomes more visible in very small configurations.}

% \textbf{Impact of Network Size ($N_{\text{src}}$):} To evaluate scalability, we compare SNN/QNN efficiency for a small network ($N_{\text{src}}=64$) versus a larger network ($N_{\text{src}}=4096$), using both 4-bit and 8-bit weights and varying $T$ and $\SR$ (Figure~\ref{fig:sensitivity_detail_plots}). While absolute energy values scale significantly with $N_{\text{src}}$ (as most dominant energy terms are proportional to $N_{\text{src}}$), the relative efficiency trends and the $\SR$ break-even points do not drastically change. For example, for $T=5$ and 4-bit weights, the SNN-QNN breakeven $\SR$ is $ 0.058$ for $N_{\text{src}}=4096$ and remains similar at $ 0.057$ for $N_{\text{src}}=64$. This suggests that while $N_{\text{src}}$ dictates the magnitude of energy consumption, its impact on the \textit{relative} SNN vs. QNN efficiency ranking (for a given $T, \SR$) is much less pronounced than that of $T$ or $\SR$ itself.

\textbf{Impact of Weight Precision:} Increasing weight precision relaxes the sparsity requirement for an SNN to achieve energy parity with a QNN, allowing the SNN to remain competitive at higher spike rates ($\SR$). For instance, in our evaluation with $T=2$ and $N_{\text{src}}=4096$, the breakeven spike rate $\SR$ increases from $0.104$ for 4-bit weights to $ 0.123$ for 8-bit weights. This trend is because higher precision disproportionately increases the energy of a QNN's MAC operations relative to an SNN's ACCs, amplifying the SNN's efficiency. While both architectures consume more power at higher precision, the QNN's baseline energy rises more sharply than the SNN's energy. This dominant effect shifts the energy crossover point, making the SNN advantageous over a wider operational range.

\revisev{\textbf{Overall Insights:} The time window and spike rate remain the dominant algorithmic parameters governing relative energy efficiency. Under the Typical Neuromorphic setting, with 4-bit weights and $N_{\text{src}}=4096$, the updated $T=5$ breakeven remains approximately 5.7\% after including membrane-state accesses. Larger $T$ generally tightens this requirement, emphasizing the need to co-design SNNs for both short time windows and low activity.}

%\textbf{Overall Insights:} Our sensitivity analysis identifies the SNN time windowsize and spike rate as the most critical algorithmic parameters governing relative energy efficiency. Our results quantify a clear trade-off: as $T$ increases, the permissible spike rate for an SNN to remain energy-competitive shrinks dramatically. Under typical neuromorphic settings, our analysis shows that for SNNs to be more efficient than a comparable QNN, the spike rate ($\SR$) must be kept below 5.7\% once the time window ($T$) exceeds five timesteps. This highlights the paramount importance of co-designing SNN models to operate with both short time windows and extreme sparsity.

\subsection{Analyses on the effect of mapping}
\label{sec:eval-data-movement-energy}
\noindent\textbf{\revise{Routing distance}}
In previous setting, we assumed adjacent layers are placed on adjacent cores, implying a hop count of 1. 
However, to provide a more comprehensive and realistic analysis, we relax the fixed hop count assumption by introducing $k_{\text{hop}}$, a factor representing the average number of inter-core hops per \revise{spike} transfer. 
\revise{This lets us study mapping scenarios from the best case with fully local activations ($k_{\text{hop}}=0$) to practical low-hop deployments. For example, SNEAP evaluates SNN mapping on a 5$\times$5 2D-mesh NoC with crossbar-based cores and uses an average hop count of 1.5 as a mapping metric~\cite{10.1145/3386263.3406900}. We also use a VGG16-related reference based on neuromorphic mesh simulation with CanMore, which adopts mesh topology with XY routing and reports an average hop count of 0.64~\cite{10247850,10378556}.
We incorporated this $k_{\text{hop}}$ factor by scaling the base data movement costs, $\Emovesparsecustom$ and $\Emovedensecustom$, showing in Equation~(\ref{eq:ann_data_sparse_hop}-\ref{eq:snn_data_dense_hop}). 
For mesh-based systems with shortest-path, Manhattan distance captures hop count and therefore serves as one proxy for mapping quality~\cite{10.1145/3582016.3582038}.}

\begin{figure*}[t]
    \centering
    \setlength{\tabcolsep}{2pt}
    \captionsetup[subfigure]{justification=centering}

    \begin{subfigure}[t]{0.24\textwidth}
        \centering
        \includegraphics[width=\linewidth,clip,trim=8 8 8 8]{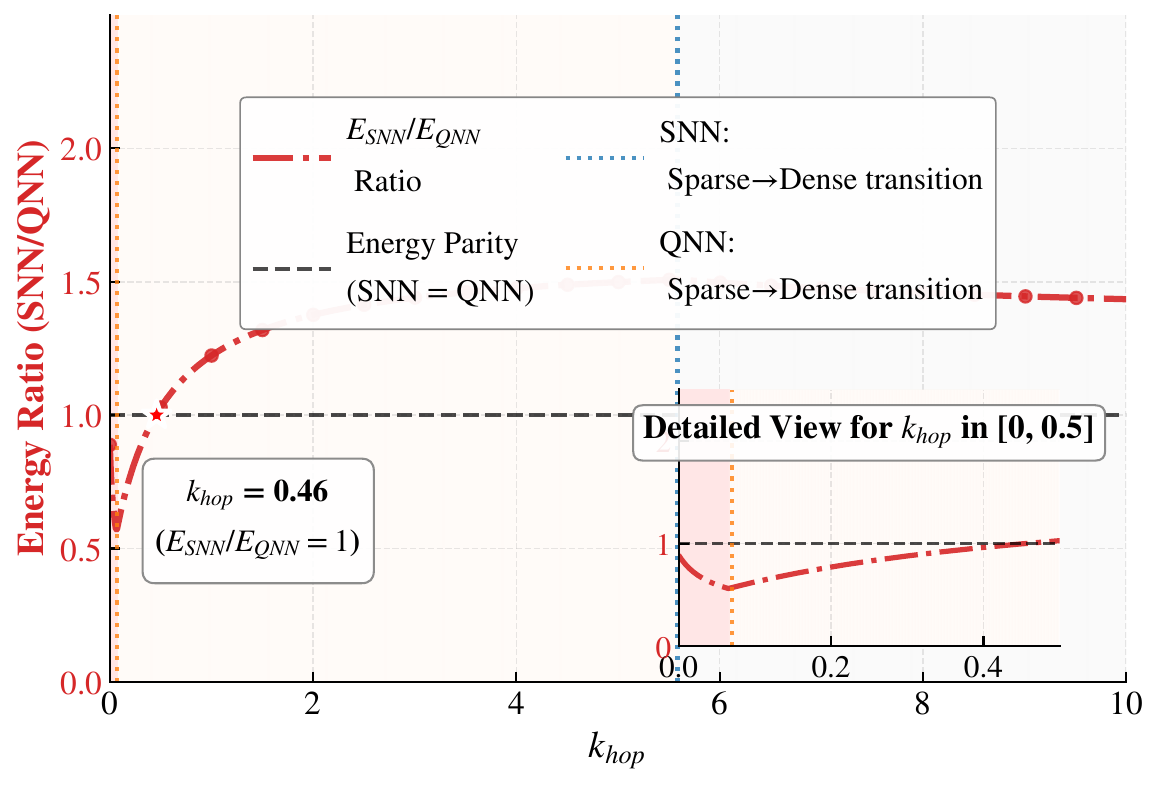}
        \caption{$R_{\mathrm{QNN}}, R_{\mathrm{SNN}} = 1, 1$}
        \label{fig:hop_r1_r1}
    \end{subfigure}
    \hfill
    \begin{subfigure}[t]{0.24\textwidth}
        \centering
        \includegraphics[width=\linewidth,clip,trim=8 8 8 8]{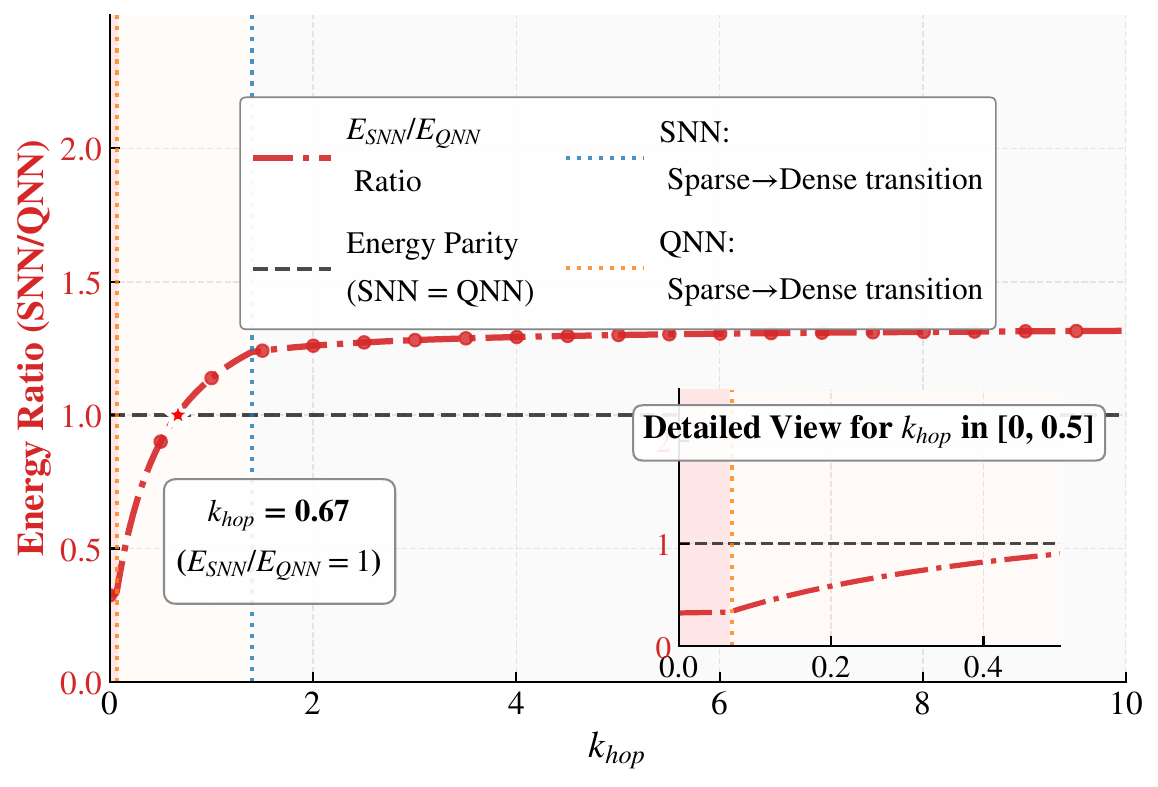}
        \caption{$R_{\mathrm{QNN}}, R_{\mathrm{SNN}} = 1, T$}
        \label{fig:hop_r1_r4}
    \end{subfigure}
    \hfill
    \begin{subfigure}[t]{0.24\textwidth}
        \centering
        \includegraphics[width=\linewidth,clip,trim=8 8 8 8]{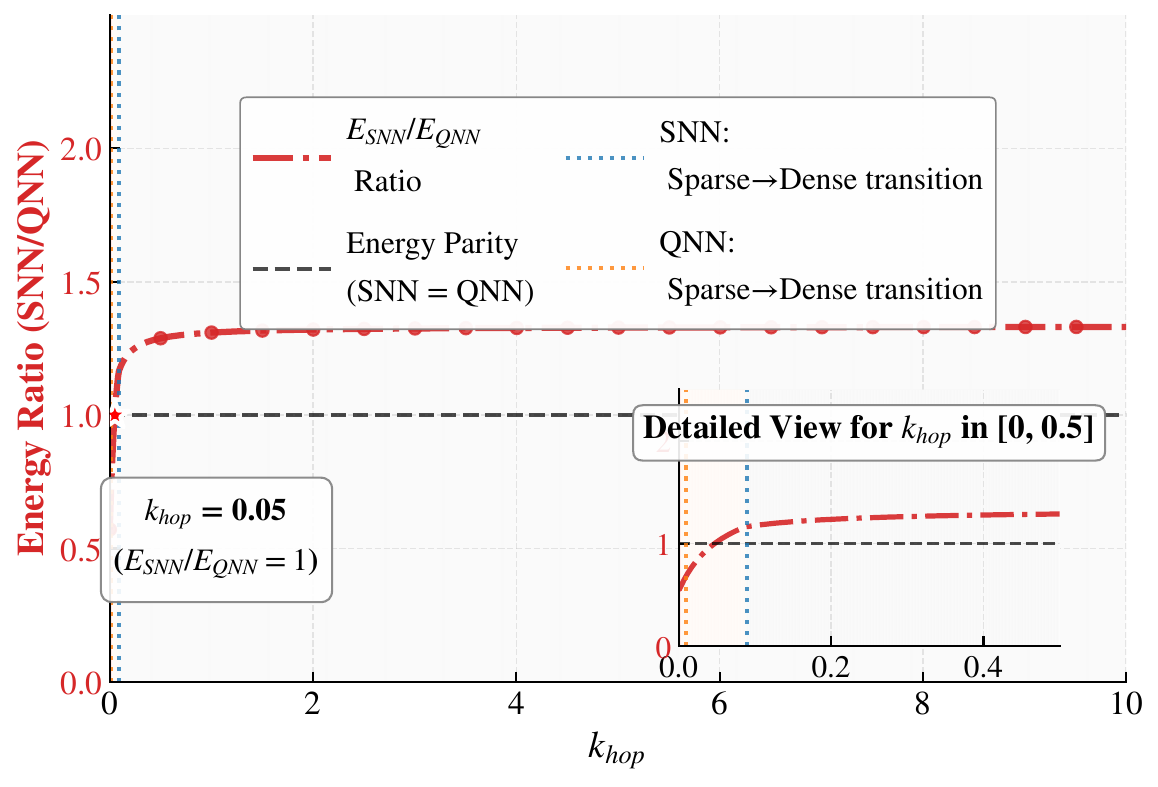}
        \caption{$R_{\mathrm{QNN}}, R_{\mathrm{SNN}} = 64, 64$}
        \label{fig:hop_r64_r64}
    \end{subfigure}
    \hfill
    \begin{subfigure}[t]{0.24\textwidth}
        \centering
        \includegraphics[width=\linewidth,clip,trim=8 8 8 8]{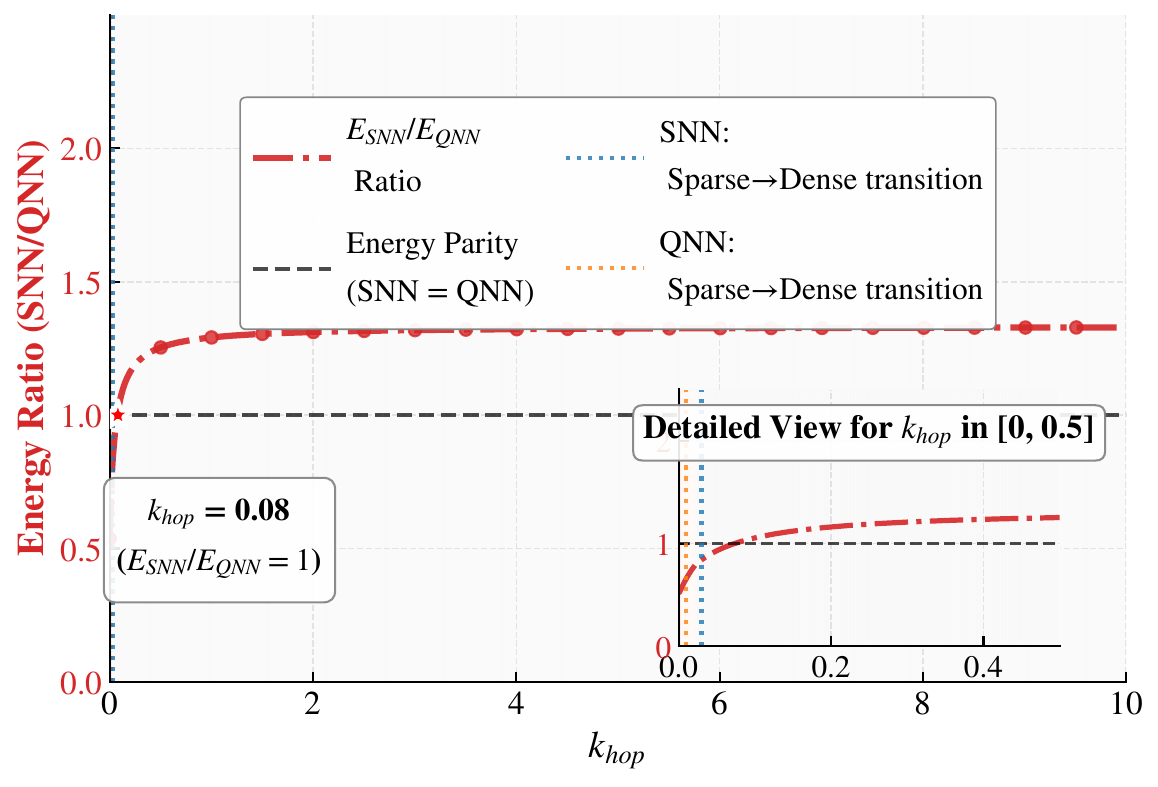}
        \caption{$R_{\mathrm{QNN}}, R_{\mathrm{SNN}} = 64, 64*T$}
        \label{fig:hop_r64_r256}
    \end{subfigure}

    \caption{Ablation study on the number of hops under different reuse settings.}
    \label{fig:hop_ablation_fourpanel}
\end{figure*}

% \begin{figure*}[htbp]
%     \centering
%     \includegraphics[width=1.0\linewidth]{figures/khop.pdf}
%     \caption{ Ablation studys for number of hops}
%     \label{fig:khop}
% \end{figure*}

% Put this in your preamble:
% \usepackage{graphicx}
% \usepackage{subcaption}

\begin{figure*}[t]
    \centering
    \captionsetup[subfigure]{justification=centering}

    \begin{subfigure}[t]{0.3\textwidth}
        \centering
        \includegraphics[width=\linewidth,clip,trim=8 8 8 8]{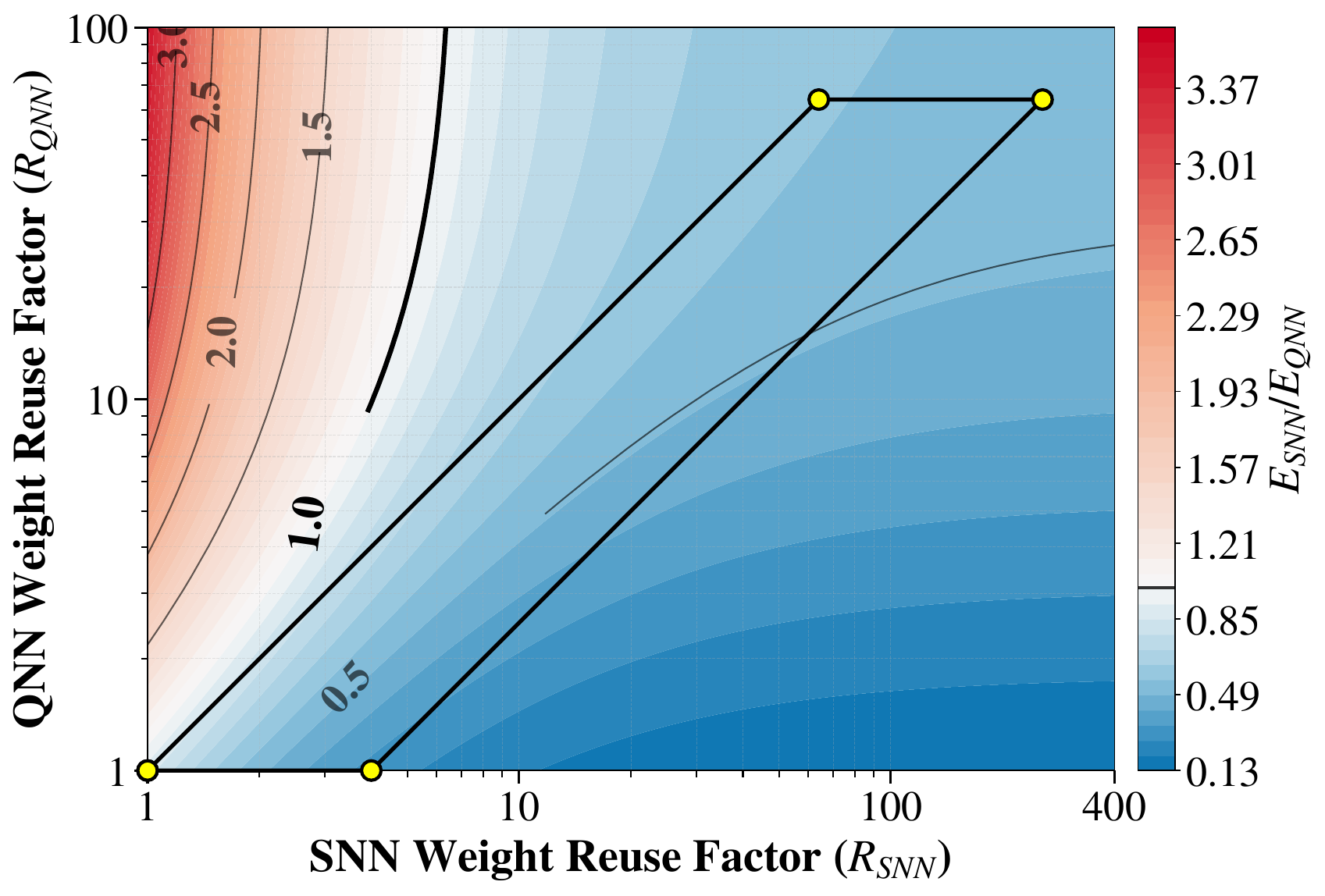}
        \caption{$k_{\mathrm{hop}}=0$}
        \label{fig:reuse_hop_0}
    \end{subfigure}
    \hfill
    \begin{subfigure}[t]{0.3\textwidth}
        \centering
        \includegraphics[width=\linewidth,clip,trim=8 8 8 8]{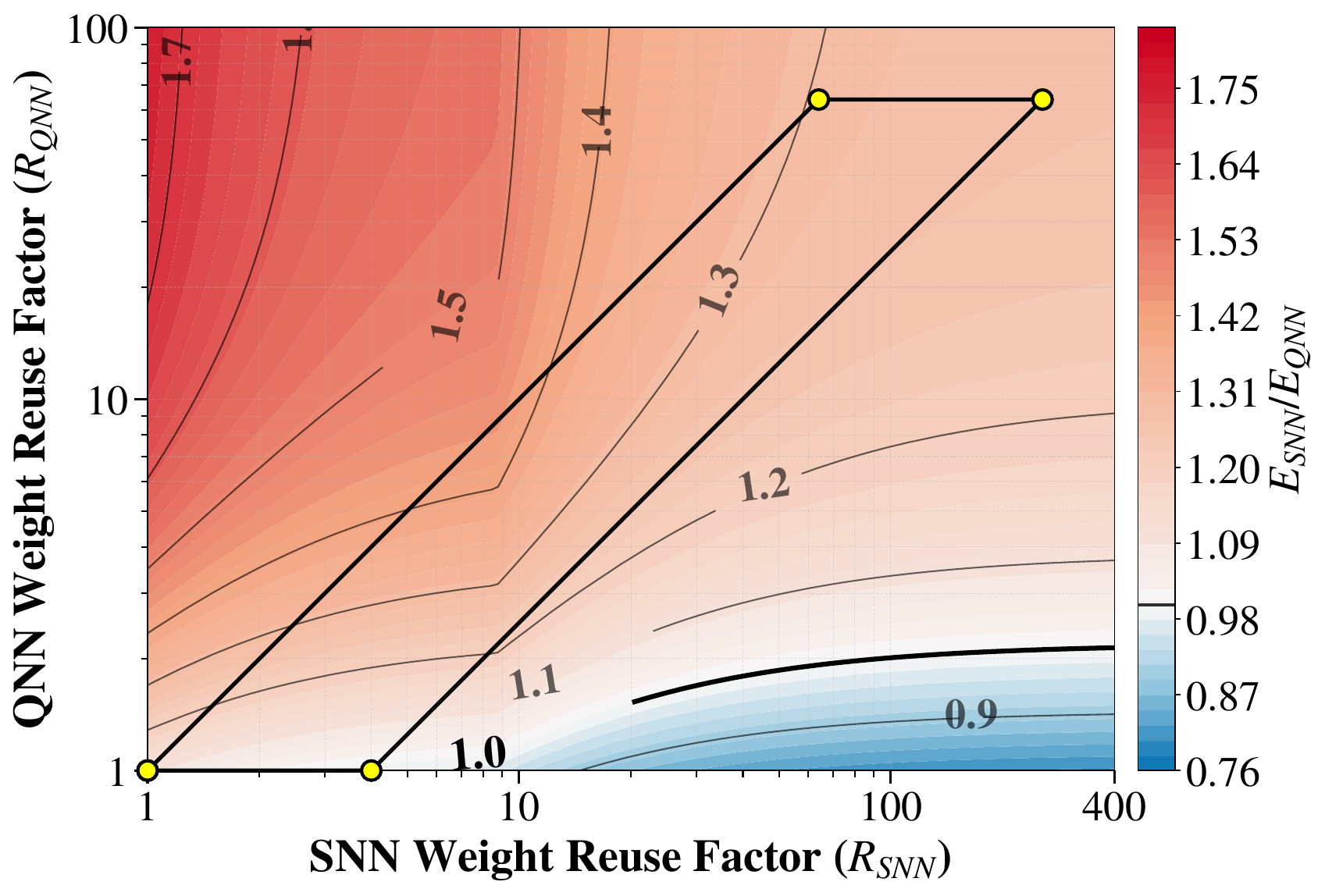}
        \caption{$k_{\mathrm{hop}}=0.64$~\cite{10378556}}
        \label{fig:reuse_hop_064}
    \end{subfigure}
    \hfill
    \begin{subfigure}[t]{0.3\textwidth}
        \centering
        \includegraphics[width=\linewidth,clip,trim=8 8 8 8]{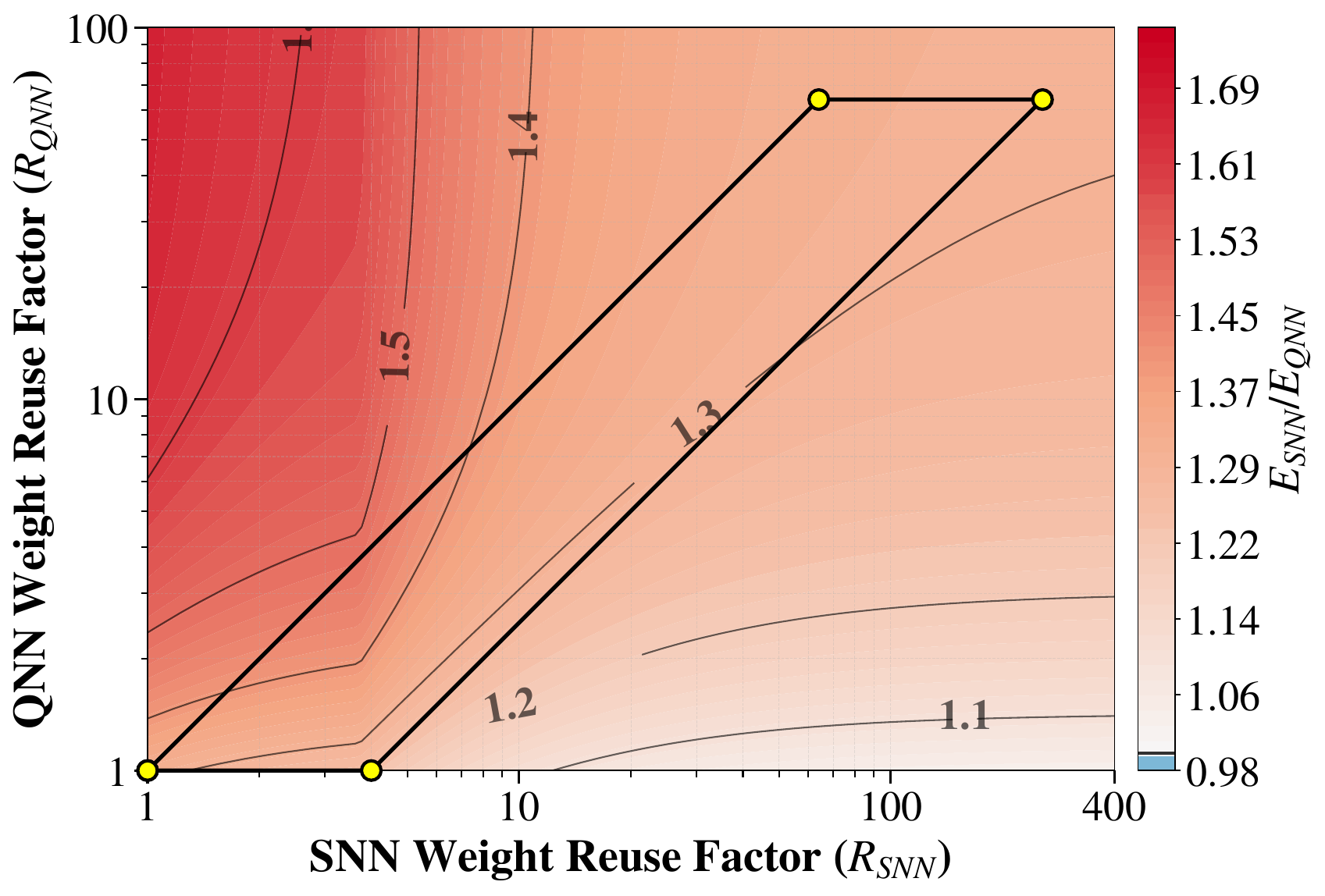}
        \caption{$k_{\mathrm{hop}}=1.5$~\cite{10.1145/3386263.3406900}}
        \label{fig:reuse_hop_15}
    \end{subfigure}

    \caption{Weight reuse factor 2D heatmaps. Each subplot (a--c) corresponds to a fixed routing cost, $k_{\text{hop}}$, of 0, 0.64, and 1.5. A ratio below 1.0 (blue) indicates that the SNN is more energy-efficient, while a ratio above 1.0 (red) favors the QNN. \revise{The parallelogram region defines the feasible ratio of $R_{\mathrm{QNN}}$ and $R_{\mathrm{SNN}}$}.}
    \label{fig:reuse_hops_threepanel}
\end{figure*}
% \begin{figure*}[htbp]
%     \centering
%     \includegraphics[width=1.0\linewidth]{figures/weight_reuse.pdf}
%     \caption{Weight reuse factors 2D heatmap. Each subplot (a-d) corresponds to a fixed routing cost, $k_{\text{hop}}$, ranging from 0.1 to 1. A ratio below 1.0 (blue) indicates the SNN is more energy-efficient, while a ratio above 1.0 (red) favors the QNN. \revise{The four marked corners of the black quadrilateral define the practical region for SNN–QNN energy comparison. $(1,1)$ denotes single-sample inference with no weight reuse. $(1,T)$ keeps a single sample but enables SNN temporal reuse over $T$ steps. $(64,64)$ denotes batch-64 inference without temporal reuse, and $(64,64T)$ combines batch-64 inference with full SNN temporal reuse. Hence, the top edge varies batch size under fixed temporal reuse, while each vertical edge isolates the effect of SNN temporal reuse at a fixed batch size. All practical deployments fall within this region.}}
%     \label{fig:weightreuse}
% \end{figure*}

\noindent\textbf{\revise{Weight reuse}}
\revise{The degree of weight reuse is highly dependent on the mapping/dataflow, the hardware system configuration, and model parameters (e.g., hidden dimension, batch size, and sequence length). To capture this effect without committing to a compiler- or hardware-specific reuse scheme, we introduce two \emph{effective weight-reuse factors}, $R_{\mathrm{QNN}}$ and $R_{\mathrm{SNN}}$, for the QNN and SNN twins, respectively. Operationally, $R$ denotes the average number of times a fetched weight can be reused before it must be loaded again, so the effective weight-loading energy is reduced from $E_{\mathrm{weight}}$ to $E_{\mathrm{weight}}/R$.

\begin{align}
\widetilde{E}^d_{\text{QNN}}
&= B S \Nsrc (1-\Egamma)
\Big(
\underbrace{\lceil\log_2(\Et+1)\rceil \, k_{\text{hop}} \, \Emovesparsecustom}_{\text{activation}}
+ \frac{E^{\text{weight}}}{R_{\text{QNN}}}
\Big)
\label{eq:ann_data_sparse_hop}
\\
\overline{E}^d_{\text{QNN}}
&= B S \Nsrc
\Big(
\lceil\log_2(\Et+1)\rceil \, k_{\text{hop}} \, \Emovedensecustom
+ \frac{E^{\text{weight}}}{R_{\text{QNN}}}
\Big)
\label{eq:ann_data_dense_}
\end{align}

\begin{align}
\widetilde{E}^d_{\text{SNN}}
&= B S \Nsrc \Et \Es
\Big(
k_{\text{hop}} \, \Emovesparsecustom
+ \frac{E^{\text{weight}}}{R_{\text{SNN}}}
\Big)
\label{eq:snn_data_sparse_}
\\
\overline{E}^d_{\text{SNN}}
&= B S \Nsrc \Et
\Big(
k_{\text{hop}} \, \Emovedensecustom
+ \frac{E^{\text{weight}}}{R_{\text{SNN}}}
\Big)
\label{eq:snn_data_dense_hop}
\end{align}

Importantly, $R_{\mathrm{QNN}}$ and $R_{\mathrm{SNN}}$ are \emph{not} assumed to be equal in general. $R_{\mathrm{QNN}}$ captures the reuse available to the QNN under a given mapping (e.g., spatial reuse, batching, or sequence-level reuse), whereas $R_{\mathrm{SNN}}$ captures the corresponding reuse for the SNN and may additionally include \emph{temporal reuse across timesteps}. Since the same SNN weight can participate in computation over $T$ timesteps, a hardware design that keeps weights resident near compute units can make $R_{\mathrm{SNN}}$ significantly larger than $R_{\mathrm{QNN}}$, potentially by as much as a factor of $T$, and in this case, the total weight loading of SNN and QNN is the same. In contrast, if weights cannot be retained across timesteps and must be reloaded at each step, then the SNN does not benefit from this temporal reuse and $R_{\mathrm{SNN}}$ becomes comparable to $R_{\mathrm{QNN}}$.
This distinction is hardware-dependent. For example, previous SNN designs~\cite{yan2024sparrowsnn,lee2021accurate,lee2022parallel,yin2024loas} show that temporal reuse across timesteps can be enabled by buffering intermediate temporal states and deferring spike generation, but such support is not universally available. Therefore, $R_{\mathrm{QNN}}$ and $R_{\mathrm{SNN}}$ should be treated as independent hardware/mapping parameters in the general model.}

\revise{To account for routing complexity and weight reuse, we rewrite the data-movement equations and extend the previous single-batch setting to the multi-batch case. Here, $B$ and $S$ denote batch size and sequence length, respectively. For non-Transformer workloads, we set $S=1$.    
%\footnote{\revise{In the previous analysis, we omitted the factor $B \cdot S$ for modeling a single-batch, non-trasformer workload. Now including $B$ and $S$ would not change the energy ratio $E_{\mathrm{SNN}}/E_{\mathrm{QNN}}$.}}
}

% \revise{Considering these two factors—routing complexity and weight reuse—we now rewrite the data movement equations and expand the previous single-batch case into multiple-batch~\footnote{\revise{B and S indicates batch size and sequence length. In non-transformer type, S is set as 1}} as follows:}

%\revise{To account for routing complexity and weight reuse, we rewrite the data-movement equations and extend the previous single-batch setting to the multi-batch case. $B$ and $S$ denote batch size and sequence length, respectively. For non-Transformer workloads, we set $S=1$~\footnote{In previous case, we didn't include $B \cdot S$ because without considering reuse factor, such $B \cdot S$ can be约掉when computing the energy radio of SNN/QNN}:}

\revise{\textbf{1) Analysis results on the number of hops $k_{\mathrm{hop}}$:}
Figure~\ref{fig:hop_ablation_fourpanel} shows the SNN/QNN energy ratio for a typical SNN setting ($T=4$, $s_r=0.1$) as a function of $k_{\mathrm{hop}}$ under four representative reuse settings: $(R_{\mathrm{QNN}},R_{\mathrm{SNN}})=(1,1)$, $(1,T)$, $(64,64)$, and $(64,64T)$. These cases cover settings from single-sample inference ($R_{\mathrm{QNN}} = 1$) without temporal reuse $(R_{\mathrm{SNN}} = R_{\mathrm{QNN}})$ to batch-64 inference ($R_{\mathrm{QNN}} = 1$) with full SNN temporal reuse $(R_{\mathrm{SNN}} = R_{\mathrm{QNN}})$. For $(R_{\mathrm{QNN}},R_{\mathrm{SNN}})=(1,1)$, i.e. single-batch, SNN without temporal reuse, the SNN/QNN ratio first drops to about 0.5 and then rises as $k_{\mathrm{hop}}$ increases. }This suggests that SNNs benefit more from sparse activations because they can exploit bit-level sparsity, while QNNs are relatively better for dense data movement. The same overall trend appears in all four cases: larger $k_{\mathrm{hop}}$ hurts the SNN relative to its twin QNN. When $k_{\mathrm{hop}}$ is small, weight loading is still an important part of the total cost, and the SNN benefits from sparse activations. As $k_{\mathrm{hop}}$ grows, activation movement becomes the main cost, both models favor dense movement, and the energy ratio increases. The parity point also depends strongly on reuse. \revise{Without reuse, parity occurs at $k_{\mathrm{hop}}\approx 0.46$. With only SNN temporal reuse, it shifts to $k_{\mathrm{hop}}\approx 0.67$. When both models already have strong reuse, the parity point drops to about $0.05$, and with additional SNN temporal reuse, it increases only slightly to about $0.08$.}

\revise{\textbf{2) Analysis results on the reuse factor:}
Figure~\ref{fig:reuse_hops_threepanel} shows the SNN/QNN energy ratio as a joint function of $R_{\mathrm{QNN}}$ and $R_{\mathrm{SNN}}$ under three routing settings, $k_{\mathrm{hop}}\in\{0,0.64,1.5\}$ where the representative $0.64$ and $1.5$ are reported by implementation on real digital neuromorphic devices~\cite{10378556, 10.1145/3386263.3406900}. The color represents how SNN is compared to the twin-QNN. Toward dark blue is where SNN is advantageous and dark red is where QNN wins. The parallelogram shows the feasible ratio of SNN and QNN's reuse factor, $(1,1)$ for single-sample inference with no weight reuse. $(T,1)$ keeps a single sample but enables SNN temporal reuse over $T$ steps. $(64,64)$ for batch-64 inference without temporal reuse, and $(64T,64)$ combines batch-64 inference with full SNN temporal reuse. }  \revise{As we can see from Figure ~\ref{fig:reuse_hops_threepanel}, }routing cost strongly affects the size of this region. At low $k_{\mathrm{hop}}$ (e.g., 0 or 0.64), SNNs have a wider energy-efficient region. The reason is that, at small $k_{\mathrm{hop}}$, activation movement is cheap compared with weight loading. In this regime, both SNNs and QNNs prefer sparse mode because sparsity reduces the weight-loading cost, but SNNs benefit more because they can further exploit bit-level sparsity. In contrast, when $k_{\mathrm{hop}}$ becomes large, such as 1.5, weight loading becomes a small part of the total cost, and activation movement dominates. Both models then favor dense mode, where the SNN no longer benefits much from sparsity and therefore has a harder time outperforming its twin QNN.

\revise{\subsection{\revisev{Trace-Driven Evaluation}}
\label{sec: real_map}
\revisev{To instantiate the analytical model with workload-dependent operating
points, we evaluate capacity-matched QNN--SNN pairs on CV and Transformer
workloads over CIFAR-10/100, ImageNet, and GLUE. From these executions,
we directly measure the QNN activation density $1-\gamma$ and the SNN
spike rate $s_r$, and substitute them into the independently defined
hardware energy model.
This trace-driven study does not serve as external validation of the
hardware model. Instead, it determines where each workload lies within
the analytical break-even design space. Across the evaluated cases, the
accuracy gaps between the twins remain small, supporting their use for a
capacity-matched energy comparison.}}

\textbf{VGG16 across time windows.}
For this analysis, we set the average hop count to $k_{\mathrm{hop}}=0.64$~\cite{10378556}.
%\footnote{\revise{A VGG16-related reference from a neuromorphic mesh simulation using CanMore~\cite{10247850} reports 71M spikes/frame and 45.4M hops/frame for VGG16, which implies about 0.64 hops per spike.}}
We set $R_{\mathrm{QNN}}=B \cdot H_{\mathrm{out}} \cdot W_{\mathrm{out}}$ to model full spatial reuse of each kernel, and $R_{\mathrm{SNN}}=B \cdot T \cdot H_{\mathrm{out}} \cdot W_{\mathrm{out}}$ to capture the extra temporal reuse across $T$ steps. Table~\ref{tab:snn_qnn_comparison} shows that QNN--SNN conversion is nearly lossless on both CIFAR-10 and CIFAR-100. In contrast, the energy trend becomes less favorable to SNNs as $T$ increases. On CIFAR-10, the SNN is slightly better at $T=3$ ($E_{\mathrm{SNN}}/E_{\mathrm{QNN}}=\revisev{0.987}$), and becomes clearly worse at larger $T$ ($\revisev{1.406}$ to $\revisev{2.243}$ for $T=5$ to $7$). On CIFAR-100, the SNN is already worse at $T=3$ ($\revisev{1.262}$), and the gap grows further as $T$ increases.

 \revisev{  \textbf{Weight sparsity}: we also use the
  bitmap-based sparse DNN encoding of
 Wang et al~\cite{9499745}.
  Let $N_w$ be the number of original weight positions, $B_w$ the weight bitwidth,
  and $\rho_w$ the zero-weight fraction. The format uses one bitmap bit per
  original position and stores only the nonzero $B_w$-bit values, giving
  \[
  S_{\mathrm{bitmap}}
  =N_w+B_wN_w(1-\rho_w),
  \qquad
  S_{\mathrm{dense}}=B_wN_w.
  \]
  Allowing each twin to select the lower-traffic dense or bitmap format gives
  \[
  f_{\mathrm{move}}(\rho_w,B_w)
  =\min\!\left\{1,(1-\rho_w)+\frac{1}{B_w}\right\}.
  \]
  
   Across $\rho_w\in[0,0.9]$, the maximum relative change in
  $E_{\mathrm{SNN}}/E_{\mathrm{QNN}}$ is only $0.10\%$ with bitmap-compressed
  weight traffic and $2.82\%$ when both twins additionally apply ideal
  zero-weight synaptic-compute gating. Neither case changes which twin has
  lower energy in any of the 12 configurations.
}

\begin{table}[htb]
\centering
\caption{QNN vs.\ SNN accuracy and sparsity for VGG16 on CIFAR10 and CIFAR100 across time steps $T$.}
\label{tab:snn_qnn_comparison}
\small
\resizebox{1\linewidth}{!}{
\begin{tabular}{lcccccc}
\toprule
& \multicolumn{6}{c}{\textbf{Time Steps ($T$)}} \\
\cmidrule(lr){2-7}
\textbf{Metric } & \textbf{3} & \textbf{4} & \textbf{5} & \textbf{6} & \textbf{7} & \textbf{8} \\
\midrule
\multicolumn{7}{l}{\textbf{CIFAR100 / VGG16}}\\
QNN accuracy(\%)                & 63.31 & 69.91 & 71.52 & 72.16 & 72.26 & 72.80 \\
SNN accuracy(\%)                & 63.24 & 69.72 & 71.43 & 72.08 & 72.45 & 72.97 \\
QNN avg.\ density rate (\%)     & 15.39 & 21.52 & 26.34 & 29.45 & 32.30 & 34.30 \\
SNN avg.\ spike rate (\%)       & 7.09  & 8.39  & 9.52  & 10.04 & 10.69 & 10.94 \\
SNN / QNN energy                & \revisev{1.262} & \revisev{1.316} & \revisev{1.635} & \revisev{1.953} & \revisev{2.270} & \revisev{1.958} \\
\midrule
\multicolumn{7}{l}{\textbf{CIFAR10 / VGG16}}\\
QNN accuracy (\%)               & 81.73 & 90.86 & 92.16 & 93.01 & 93.09 & 92.93 \\
SNN accuracy (\%)               & 82.09 & 90.71 & 91.82 & 92.93 & 93.11 & 92.94 \\
QNN avg.\ density rate (\%)     & 12.20 & 16.52 & 19.89 & 22.98 & 25.45 & 27.39 \\
SNN avg.\ spike rate(\%)        & 5.49  & 6.43  & 7.10  & 7.73  & 8.19  & 8.52 \\
SNN / QNN energy                & \revisev{\textbf{0.987}} & \revisev{1.026} & \revisev{1.406} & \revisev{1.824} & \revisev{2.243} & \revisev{1.966} \\
\bottomrule
\end{tabular}}
\end{table}

%\revise{\textbf{Generalization to other CV models.}
%Table~\ref{tab:cv_benchmarks} shows the same trend beyond VGG16. Under low-$T$ settings (T=3), the SNN can remain close to the QNN and may even be slightly better, as in VGG16 on CIFAR-10 ($0.982$). However, this advantage is not stable across models or datasets. ResNet-18 gives ratios of $1.065$ on CIFAR-10 and $1.017$ on CIFAR-100, while RepVGG~\cite{ding2021repvgg} on ImageNet reaches $2.228$. For all CV models, we use the same hop count and reuse setting as in VGG16. These results show that SNNs are only competitive in a narrow regime with short time windows and low spike rates. As activity or $T$ increases, the QNN becomes more energy efficient.

\revise{\textbf{Generalization to other CV models.}
Table~\ref{tab:cv_benchmarks} shows the same trend beyond VGG16. Under low-$T$ settings ($T=3$), the SNN can remain close to the QNN and may even be slightly better, as in VGG16 on CIFAR-10 ($\revisev{0.987}$). However, this advantage is not stable across models or datasets. ResNet-18 gives ratios of $\revisev{1.071}$ on CIFAR-10 and $\revisev{1.023}$ on CIFAR-100, while RepVGG~\cite{ding2021repvgg} on ImageNet reaches $\revisev{2.232}$. For all CV models, we use the same hop count and reuse setting as in VGG16. These results show that SNNs are only competitive in a narrow regime with short time windows and low spike rates. As activity or $T$ increases, the QNN becomes more energy efficient.
The slight non-monotonicity at (T=8) arises because the QNN bit width increases from 3 to 4 bits.
}
\begin{table}[t]
\centering
\caption{\revise{CV benchmarks. We use $T=3$ for VGG16 and ResNet-18, and $T=10$ for RepVGG (RepVGGplus-L2pse) on ImageNet, which requires a slightly larger time window to reach comparable accuracy~\cite{yan2022low}.}}
\label{tab:cv_benchmarks}
\setlength{\tabcolsep}{6pt}
\resizebox{1\linewidth}{!}{
\revise{
\begin{tabular}{lccccc}
\toprule
 & \multicolumn{2}{c}{\textbf{CIFAR-10}} & \multicolumn{2}{c}{\textbf{CIFAR-100}} & \textbf{ImageNet} \\
\cmidrule(lr){2-3} \cmidrule(lr){4-5} \cmidrule(lr){6-6}
 & \textbf{VGG16} & \textbf{ResNet-18} & \textbf{VGG16} & \textbf{ResNet-18} & \textbf{RepVGG} \\
\midrule
QNN accuracy (\%)     & 81.73 & 91.68 & 63.31 & \revisev{69.22} & 78.274 \\
SNN accuracy (\%)     & 82.09 & 91.75 & 63.24 & 67.52 & 78.232 \\
QNN density rate (\%) & 12.20 & 9.10  & 15.39 & 8.79  & 32.88 \\
SNN spike rate (\%)   & 5.49  & 5.90  & 7.09  & 5.63  & 12.41 \\
\textbf{SNN / QNN energy} & \revisev{\textbf{0.987}} & \revisev{\textbf{1.071}} & \revisev{\textbf{1.262}} & \revisev{\textbf{1.023}} & \revisev{\textbf{2.232}} \\
\bottomrule
\end{tabular}}}
\end{table}

\revise{\textbf{Transformer workloads.}
For Transformer workloads, we use $k_{\mathrm{hop}}=4.75$, based on the average hop count reported for transformer-like models in Allspark (range: 2.59--8.48)~\cite{10726919}. In practice, however, the final energy ratio is  insensitive to $k_{\mathrm{hop}}$. The reason is that both QNNs and SNNs fall into the dense data-movement regime for Transformer workloads because the QNN density and SNN spike rate are both relatively high. At the same time, the large reuse factors, $R_{\mathrm{QNN}}=B\cdot S$ and $R_{\mathrm{SNN}}=B\cdot S\cdot T$, make the amortized weight-loading cost negligible. Under this regime, the dominant term is the temporal factor, which gives
$
\frac{E_{\mathrm{SNN}}}{E_{\mathrm{QNN}}}
\approx
\frac{T}{\lceil \log_2(T+1) \rceil}.
$

Table~\ref{tab:sst2_across_t} shows that, on SST-2, the SNN/QNN energy ratio rises from 0.996 at $T=1$ to \revisev{1.500}, \revisev{2.338}, and \revisev{3.762} at $T=3$, 7, and 15, respectively. The same trend appears across all seven GLUE tasks at the 2-bit point ($T=3$) in Table~\ref{tab:glue_7_tasks_2bit_t3}: the energy ratio stays near \revisev{1.500}, while the accuracy gap between the twins remains small.

\begin{table}[hbt]
\centering
\caption{\revise{SST-2 across $T \in \{1,3,7,15\}$.}}
\label{tab:sst2_across_t}
\revise{
\begin{tabular}{lcccc}
\toprule
 & $T=1$ & $T=3$ & $T=7$ & $T=15$ \\
\midrule
QNN accuracy (\%) & 68.23 & 90.14 & 91.74 & 91.74 \\
SNN accuracy (\%) & 66.51 & 89.91 & 91.51 & 91.63 \\
QNN density rate (\%) & 47.54 & 53.03 & 58.27 & 59.58 \\
SNN spike rate (\%) & 43.34 & 34.01 & 31.63 & 30.36 \\
SNN / QNN energy & 0.996 & \revisev{1.500} & \revisev{2.338} & \revisev{3.762} \\
\bottomrule
\end{tabular}}
\end{table}

\begin{table}[hbt]
\centering

\caption{\revise{GLUE 7 tasks at 2-bit with $T=3$.}}
\label{tab:glue_7_tasks_2bit_t3}
\scriptsize
\setlength{\tabcolsep}{3pt}
\revise{
\begin{tabular}{lccccccc}
\toprule
 & MNLI & QQP & QNLI & SST-2 & STS-B & RTE & MRPC \\
\midrule
QNN accuracy (\%) & 78.03 & 85.92 & 87.22 & 90.14 & 83.83 & 70.40 & 88.19 \\
SNN accuracy (\%) & 77.59 & 85.80 & 86.71 & 89.91 & 83.13 & 69.31 & 87.13 \\
QNN density rate (\%) & 50.12 & 51.19 & 48.49 & 53.03 & 49.84 & 55.71 & 48.64 \\
SNN spike rate (\%) & 32.78 & 31.42 & 31.25 & 34.01 & 32.23 & 35.87 & 31.61 \\
SNN / QNN energy & \revisev{1.501} & \revisev{1.500} & \revisev{1.500} & \revisev{1.500} & \revisev{1.500} & \revisev{1.500} & \revisev{1.501} \\
\bottomrule
\end{tabular}}
\end{table}

\revisev{The same behavior holds as models scale up. On a spiking Llama-2 7B at
$T=15$ with batch size $B=64$ and sequence length $S=128$, the measured
energy ratio is $E_{\mathrm{SNN}}/E_{\mathrm{QNN}}=3.793$, again matching
the temporal factor $T/\lceil\log_2(T{+}1)\rceil=3.75$ and confirming that
the same dense-movement regime governs larger Transformers.}
}
%Thus, the real-workload results match the analytical trend: SNNs are energy efficient only in a narrow regime with small time windows and high sparsity.}

\section{Discussion}
\label{sec:discussion}

\revise{
\subsection{Latency}
Although this work focuses on first-principles energy, the same framework also suggests a first-order latency trade-off. Relative to its twin QNN, an SNN distributes inference over
$
T
$
timesteps and executes approximately
$
N_{\mathrm{src}}\,T\,\SR
$
active accumulations, plus per-step threshold/reset operations, whereas the QNN executes approximately
$
N_{\mathrm{src}}(1-\gamma)
$
active MACs in one pass. Hence, from an operation-count perspective, an SNN can only obtain a first-order latency-side advantage when
$
T\,sr \lesssim (1-\gamma),
$
or more generally,
$
T\,sr \lesssim k(1-\gamma),
$
with
$
k = E_{\mathrm{MAC}}/E_{\mathrm{ACC}}
$.
This bound is optimistic, however, because it assumes that sparsity can be exploited with negligible control overhead. In real systems, latency is also affected by event-routing cost, arbitration, synchronization, spike-distribution-induced load imbalance, routing distance, and weight reuse. These factors depend strongly on system scale and mapping, and can create non-trivial energy--latency trade-offs. For example, while sparsity may reduce arithmetic work and energy, it may also lower arithmetic intensity and thereby hurt scalability on larger systems. For this reason, the present paper centers on energy, whose dominant first-principles components are more portable across implementations than latency.
}

\revisev{
\subsection{Relation to Prior Energy Analyses}
\label{sec:prior}

Previous studies establish that operation counts alone are insufficient to determine SNN energy efficiency. Dampfhoffer et al.~\cite{dampfhoffer2022snns} combine analytical bounds on spike sparsity and data reuse with SRAM/DRAM-aware Eyeriss~v1/v2 case studies. Bhattacharjee et al.~\cite{bhattacharjee2024snns} use the SATA~\cite{yin2022sata} digital-systolic and SpikeSim~\cite{moitra2023spikesim} analog-IMC platforms to quantify repeated timestep computation, data movement, membrane-state overhead, and crossbar non-idealities. In contrast, we adopt a capacity-matched baseline that compares a rate-encoded IF SNN with its QNN twin under identical network structure, weight precision, and mapping constraints. We then derive closed-form break-even regions as functions of timestep count, spike activity, QNN precision, sparse and dense movement costs, weight loading, and data reuse. Rather than evaluating whether a specific SNN--ANN pair is more efficient under one hardware setting, our analysis identifies where a rate-encoded SNN becomes more energy-efficient than its capacity-matched QNN twin across the joint algorithm--hardware design space.

Specialized SNN accelerators, including LoAS~\cite{yin2024loas}, SpikeX~\cite{xu2025spikex}, and Bishop~\cite{xu2025bishop}, improve efficiency partly by reducing sparse-event movement overhead. To quantify this effect, we fix $\overline{E}^{\mathrm{move}}=0.25$~pJ/bit/hop and sweep
$
\eta=
\frac{\widetilde{E}^{\mathrm{move}}}
{\overline{E}^{\mathrm{move}}}
$
while allowing each SNN and parallel-MAC QNN layer to select its lower-energy
sparse or dense movement branch. Across 12 rate-coded VGG16 configurations
($T=3,\ldots,8$ on CIFAR-10 and CIFAR-100), the SNN/QNN energy ratio is
$0.594$--$0.872$ at $\eta=1$, with the SNN lower in all 12 configurations. At
$\eta=4$, the ratio is $0.541$--$1.020$ and the SNN is lower in 11
configurations; at the nominal $\eta=12$, it is $0.987$--$2.270$ and the SNN is
lower in only one configuration. At $\eta=64$, the ratio is
$1.316$--$2.282$, and the SNN is lower in none. These results show that reducing
sparse-event movement cost can substantially expand the region in which the SNN
is more energy-efficient than its capacity-matched QNN twin.

On the algorithm side, a complementary line of work lowers the inputs to our
model rather than the hardware cost: Temporal-Switch-Coding (TSC) cuts the
spike count through time-based coding~\cite{han2020deep}; STDP-based
pruning with weight quantization~\cite{rathi2018stdp} and attention-guided
compression in Spike-Thrift~\cite{kundu2021spike} reduce weight density and
firing activity; and high-order information-bottleneck training yields
compact, robust SNNs for recognition~\cite{yang2023effective} and
event-based optical flow~\cite{yang2024self}. These methods lower $T$ or $s_r$ and thus improve SNN
efficiency.
}

\revisev{
\subsection{Bit-Level QNN Comparison}
\label{sec:bitlevel}

We further consider a bit-serial QNN implementation to provide a direct
comparison with the rate-encoded SNN. Following the bit-serial execution
principles of Stripes~\cite{judd2016stripes} and Neural Cache~\cite{eckert2018neural}, we
evaluate a bit-level QNN design in addition to the synthesized parallel-MAC
baseline.

Let $K\in\{0,\ldots,T\}$ denote an integer QNN activation, and let
$\mathbf{k}=(k_0,\ldots,k_{B-1})$ be its binary representation, where
$B=\lceil\log_2(T+1)\rceil$. Thus,
\begin{equation}
K=\sum_{j=0}^{B-1}2^j k_j.
\end{equation}
The binary code is processed locally, one bit position at a time. Each active
bit $k_j=1$ contributes $2^j w$ through one Shift-ACC operation, while zero
bits gate the corresponding accumulation. The nonzero activation density is $1-\gamma=\Pr(K>0)$, thus the compute energy is

\begin{equation}
E^c_{\mathrm{bQNN}}
=
N_{\mathrm{src}}(1-\gamma)
\mathbb{E}\!\left[
\lVert\mathbf{k}\rVert_0 \mid K>0
\right]
E_{\mathrm{ShiftACC}}
+2E_{\mathrm{CMP}}.
\end{equation}

%The measured activation-code distribution further captures the number of active operations. Data-movement energy is evaluated using the same adaptive framework as in the rest of the paper.

We trace the activation codes of 12 VGG16 QNNs with 8-bit
weights on CIFAR-10 and CIFAR-100. The
measured nonzero activation density ranges from $12.199\%$ to $34.305\%$, and
each nonzero activation contains $1.080$--$1.294$ active bits on average. For CIFAR-10, the resulting energy ratio
$E_{\mathrm{SNN}}/E_{\mathrm{bQNN}}$ is $0.998$ at $T=3$, $1.041$ at $T=4$,
and $1.428$--$2.280$ for $T\geq5$. For CIFAR-100, all ratios range from
$1.279$ to $2.315$. Therefore, the SNN is slightly more efficient for
CIFAR-10 at $T=3$, reaches near parity at $T=4$, and is less efficient than
the bit-serial QNN in the remaining ten cases.

}
% A
% packed $\lceil\log_2(T+1)\rceil$-bit QNN word is transferred before local
% bit-serial execution. Per target activation (the common layer aggregation factor
% is applied afterward), the sparse and dense branches are
% \[
% \begin{aligned}
% \widetilde E^d_{\mathrm{bQNN}}
% &=N_{src}(1-\gamma)\left(
% \lceil\log_2(T+1)\rceil k_{\mathrm{hop}}\widetilde E_{\mathrm{move}}
% +\frac{E^{\mathrm{weight}}}{R_{\mathrm{QNN}}}\right),\\
% \overline E^d_{\mathrm{bQNN}}
% &=N_{src}\left(
% \lceil\log_2(T+1)\rceil k_{\mathrm{hop}}\overline E_{\mathrm{move}}
% +\frac{E^{\mathrm{weight}}}{R_{\mathrm{QNN}}}\right).
% \end{aligned}
% \]
% For every layer $l$,
% \[
% E^d_{\mathrm{bQNN},l}
% =\min\!\left(
% \widetilde E^d_{\mathrm{bQNN},l},
% \overline E^d_{\mathrm{bQNN},l}
% \right).
% \]
% The weight is retained across the $\lceil\log_2(T+1)\rceil$ local binary positions, and
% $R_{\mathrm{QNN}}$ captures mapping-level reuse. The same layer-wise selection
% rule is applied to the SNN using its $Ts_r$ sparse and $T$ dense branches.
\revisev{
\subsection{Other encoding methods}
 We discuss how other coding schemes,
  such as temporal and burst coding, can be evaluated using the same energy model in this section.
  For each scheme, we directly measure the spike rate from its spike traces and
  substitute the measured value into the model.

  As an example, we study a discrete TTFS coding scheme. Zero activations remain
  silent, while each nonzero activation generates exactly one spike within the
  $T$-step window. The resulting code provides $T+1$ distinct activation levels. We substitute the measured spike rate into the original compute and
  data-movement models. We retain the fixed $T$-step neuron-state evaluation,
  the temporal weight-reuse assumption, and the paper's original parallel-MAC
  QNN comparator. The resulting
  $E_{\mathrm{SNN}}/E_{\mathrm{QNN}}$ ratios are:

  \begin{center}
  \small
  \scalebox{0.8}{
  \begin{tabular}{lrrrrrr}
  \toprule
   & $T=3$ & $T=4$ & $T=5$ & $T=6$ & $T=7$ & $T=8$ \\
  \midrule
  CIFAR-10  & 0.732 & 0.661 & 0.790 & 0.907 & 1.001 & 0.811 \\
  CIFAR-100 & 0.915 & 0.850 & 1.031 & 1.146 & 1.255 & 1.002 \\
  \bottomrule
  \end{tabular}}
  \end{center}

  Under this sparse TTFS endpoint, the SNN is more energy-efficient in five of
  six CIFAR-10 configurations and two of six CIFAR-100 configurations.
  CIFAR-10/$T=7$ and CIFAR-100/$T=8$ are both near parity. This experiment
  isolates the effect of coding-induced spike sparsity using fixed activation
  traces and the original energy model.
}

\revisev{
\subsection{Cross-check, provenance, and sensitivity}
\textbf{Cross-check.} We use Timeloop~\cite{parashar2019timeloop} to audit the two structural
assumptions behind our action counts: that weights stay resident in local
storage, and that partial sums are not spilled and reloaded before their
reductions finish. We define $F_W$ as the average number of times each unique weight is fetched
from the parent level, and $F_P$ as the number of times partial sums are
spilled and reloaded before their reduction finishes. We audit two configurations: (i) the FC-4096 operating point behind the
headline crossover in Fig.~5 ($N_{\mathrm{src}}=4096$, $N_{\mathrm{out}}=1$,
$T=5$; one PE, a 96.08-KiB BackingStorage, a 24.02-KiB unified local Buffer,
and 8.01-KiB metadata storage); and (ii) VGG16 \texttt{conv4\_3} ($N=1$,
$C=M=512$, $R=S=3$, $P=Q=4$) on 64 PEs with a 384-KiB SharedBuffer and, per
PE, a 36-KiB WeightBuffer, an 18-KiB ActivationBuffer, and a 128-entry
16-bit Accumulator; each PE holds eight output channels, so its full
$8\times512\times3\times3=36{,}864$-weight partition fits locally. For each QNN configuration we run two independent EDP-directed random mapper
searches, keep the minimum-EDP legal mapping, and replay it with \texttt{timeloop-model}; both searches return the
same normalized mapping. Timeloop cannot model the IF recurrence directly, so we keep the frozen QNN mapping unchanged and wrap it in an outer loop that runs $T=5$ times. Since the loop is outside the weight tile, so weights are loaded once and reused across all timesteps. 

In both configurations, Timeloop reports exactly one fill per
unique weight and no extra partial-sum traffic, i.e., $F_W=1$ and $F_P=0$,
matching the model's assumptions with zero deviation. Sparseloop further
confirms on the same mappings that sparsity is applied to the intended
data-movement and compute actions. The cross-check thus confirms the validity of our energy equations for the audited configurations.

\textbf{Provenance:} We further  provide the per-op constants’ provenance in the same section. The compute constants ($E_{\mathrm{ACC}}$,
$E_{\mathrm{MAC}}$, $E_{\mathrm{CMP}}$, $E_{\mathrm{SUB}}$) come from
per-operator synthesis on a commercial 22-nm standard-cell library
(Design Compiler, TT, 0.80\,V, 25\,$^\circ$C, 1\,GHz, vectorless activity);
$E_{\mathrm{weight}}$ comes from a memory-compiler characterization of a
low-power single-port SRAM macro at the same corner. Synthesis scripts and
power reports are included in the artifact. 
 $\widetilde{E}_{\mathrm{move}}=3.0$
\,pJ/bit/hop is normalized from Loihi's reported per-event routing cost; $\overline{E}_{\mathrm{move}}=0.25$\,pJ/bit/hop is our in-house
measurement of a circuit-switched NoC at 22\,nm.

\textbf{Sensitivity:}
We test how stable the $5.7\%$ crossover is by keeping the mapping and
action counts fixed and changing
$\widetilde{E}_{\mathrm{move}}$, $\overline{E}_{\mathrm{move}}$,
$E_{\mathrm{weight}}$, $E^{S}_{\mathrm{ACC}}$, and $E^{Q}_{\mathrm{MAC}}$
one at a time by $\pm50\%$. Data movement matters most: the crossover moves
between $3.41\%$ and $10.80\%$. Weight energy shifts the crossover only to $5.39\%$--$6.08\%$, and
ACC/MAC variation moves it by at most $0.05$ percentage points. We also add an absolute $\pm1$-percentage-point error as noise in $s_r$ changes $E_{\mathrm{SNN}}/E_{\mathrm{QNN}}$ by at most 17.39\% relative
to its nominal value. So the exact
threshold is hardware-dependent, but the conclusion stands: SNNs need very
low spike activity, and the crossover is set mainly by data-movement cost.
}

\revisev{
\subsection{Effect of noise}

In this section, we discuss how leak and input noise shift the energy balance, using the same
frozen $T=5$ VGG16 checkpoints on the full CIFAR-10/100 test sets. For input noise, Gaussian noise is added in pixel space and clipped to
$[0,1]$ before normalization. At the mild $\sigma=2/255$ point, the spike
rate moves from 7.100\% to 7.162\% on CIFAR-10 and from 9.516\% to 9.598\%
on CIFAR-100, and $E_{\mathrm{SNN}}/E_{\mathrm{QNN}}$ moves from 1.406 to
1.417 and from 1.635 to 1.634. At the $\sigma=8/255$ stress point, the
spike rates become 7.780\% and 10.341\%, the ratios become 1.536 and
1.635, and accuracy falls by 9.43 and 16.82 percentage points. So noise
moves the energy ratio only a little, but it hurts accuracy much earlier
than it hurts energy. For leak, we hold weights, threshold, and $T=5$ fixed and sweep
$\lambda=0.90$--$0.99$ in steps of 0.01. As $\lambda$ grows (weaker leak),
the spike rate rises from 5.657\% to 6.569\% on CIFAR-10 and from 8.056\%
to 9.001\% on CIFAR-100, and the energy ratio rises from 1.121 to 1.301
and from 1.574 to 1.632. In other words, leak suppresses spike firing and
so lowers SNN energy.
}

\revisev{

\subsection{Limitations and Implications for compilers and training.}
We discuss two limitations: control energy and neuron-level aggregation.
For Control energy, our analysis targets specialized dataflow architectures, where control overhead is expected to be low. We do not, however, explicitly model all architecture-level control costs. Because event-driven execution may require extra scheduling and arbitration, this omission may slightly bias the comparison in favor of the SNN.
For Neuron-level aggregation: In practical networks, small neuron-level deviations — from leakage, residual membrane states, clipping, or other non-idealities — may propagate and accumulate across layers. Our network-level energy model also aggregates per-neuron statistics, so it may not fully capture layer-wise variation in spike activity. The analytical results should therefore be read as first-order estimates. 

We further provide implications for compilers and training. Compilers can use our model in two ways: place layers to reduce hop
count
(Fig.~\ref{fig:hop_ablation_fourpanel}) and keep weights on-chip across timesteps to
raise $R_{\mathrm{SNN}}$ (Fig.~\ref{fig:reuse_hops_threepanel}). Since sparse
transmission stops paying off above $s_r \approx 0.12$--$0.17$, the
compiler can also pick sparse or dense mode per layer. For training,
the targets are simple: $T \le 5$ and a spike rate under
$\approx 5.7\%$. This can be done with rate regularization or
pruning, or sparser codes
like TTFS.
}

\section{Conclusion}

This paper addresses the conditions under which spiking neural networks achieve better energy efficiency compared to equivalent quantized ANNs. Through an analysis grounded in a fair comparison between models with equivalent information representation, we find that SNN energy efficiency is not a given, but requires a confluence of favorable algorithmic and hardware properties. At a high level, the design space for energy-efficient SNNs depends on achieving a low total number of spike events (a low $T \cdot s$ product), hardware optimized for low-cost sparse event processing (minimal $\widetilde{E}^{\text{move}}$), and a substantial MAC-to-ACC energy ratio ($E_{\text{MAC}} \gg T \cdot E_{\text{ACC}}$). Our analysis provides specific design guidance based on these factors. For instance, under typical neuromorphic hardware settings, we recommend that for an SNN to be selected, its time window should be at most five timesteps ($T<=5$) with a spike rate below approximately 5.7\%. Beyond these operational points, a well-optimized QNN is often the more energy-efficient choice. In summary, this work provides a quantitative framework to guide model selection and emphasizes the necessity of algorithm-hardware co-design for realizing the energy-saving potential of SNNs.

\section{Acknowledgment}
The authors acknowledge the use of generative AI tools, including ChatGPT, during the preparation of this manuscript and the research process. These tools were used to assist with improving the readability and clarity of the manuscript, as well as to support preliminary drafting, code development, and research-related discussions. All AI-assisted content, code, and suggestions were carefully reviewed, revised, and validated by the authors. The authors take full responsibility for the accuracy, originality, and integrity of the final manuscript.

This research is partly supported by the Ministry of Education, Singapore, under the Academic Research Fund Tier 1 (FY2024) and Tier 2 research grant T2EP20125-0020 and Tier 3 grant MOE-MOET32024-0003. This research is also partially supported by the Advanced Research and Technology Innovation Centre (ARTIC) at the National University of Singapore under Grant No. AFP-RP6.
\appendix
\section{Proof for Theorems and Propositions}
\label{sec: proof}
\subsection{Proof for Theorem~1}
\begin{proof}[Proof Outline]
The argument proceeds as follows: Lemma~1 first establishes that the membrane potential $v_i^l(t)$ of target neuron $i$ in layer $l$ remains bounded within $[0,\theta_i^l)$ under the stated input condition. This ensures that the residual potential $v_i^l(T)$ at the end of the time window $T$ also lies in this range. Lemma~2 then shows that, when the influence of the bounded residual potential $v_i^l(T)$ on the total output spike count is negligible, the SNN neuron's $T+1$ discrete output levels can be matched by a QNN neuron employing at most $\lceil \log_2(T+1)\rceil$ bits. Finally, Corollary~1 addresses the non-ideal case where this simplifying assumption does not fully hold. In such cases, the effective information representation capability of the SNN neuron does not exceed, and may be lower than, that of the corresponding $\lceil \log_2(T+1)\rceil$-bit QNN neuron.

\end{proof}
\begin{lemma}
\label{lemma:bounded_potential_revised}
Consider a target IF neuron $i$ in layer $l$ without leakage and employing a reset-by-subtraction mechanism with firing threshold $\theta_i^l$. 
If at each timestep $t \in [1,T]$, the net input current
$
I_i^l(t) = \sum_{j \in \mathcal{S}_i^{l-1}} w_{ij}^l s_j^{l-1}(t)
$
satisfies $0 \le I_i^l(t) < \theta_i^l$, then the membrane potential $v_i^l(t)$ remains bounded such that
$
0 \le v_i^l(t) < \theta_i^l, \quad \forall t \in [1,T],
$
assuming $v_i^l(0)=0$.
%Consider an IF neuron without leakage and employing a reset-by-subtraction mechanism with a firing threshold $\theta^l_i$. If at each timestep $t \in [1, T]$, the net input current $I_i^l(t)$ (i.e., $\sum_j w^l_{i,j} s_j^{l-1}(t)$) satisfies $0 \leq I_i^l(t) < \theta^l_i$, then the membrane potential $v^l_i(t)$ remains bounded such that $0 \leq v^l_i(t) < \theta^l_i$ for all $t \in [1, T]$, assuming $v_i^l(0)=0$.
\end{lemma}

\begin{proof}
The membrane potential $v^l_i(t)$ evolves as $v^l_i(t) = v^l_i(t-1) + I^l_i(t) - \theta^l_i s^l_i(t)$, where $s^l_i(t) \in \{0,1\}$ is the output spike at timestep $t$, and $v_i^l(0)=0$. We prove $0 \leq v^l_i(t) < \theta^l_i$ by induction on $t$.
\textbf{Base case ($t=0$):} $v_i^l(0)=0$, so $0 \leq v_i^l(0) < \theta^l_i$ holds.
\textbf{Inductive step:} Assume $0 \leq v^l_i(t-1) < \theta^l_i$ for some $t \ge 1$.
At timestep $t$:
\begin{itemize}
    \item \textbf{No spike emitted ($s^l_i(t) = 0$):} This occurs if $v^l_i(t-1) + I^l_i(t) < \theta^l_i$. Then, $v^l_i(t) = v^l_i(t-1) + I^l_i(t)$. Since $v^l_i(t-1) \geq 0$ and $I^l_i(t) \geq 0$ , $v^l_i(t) \geq 0$. By the condition for no spike, $v^l_i(t) < \theta^l_i$. Thus, $0 \leq v^l_i(t) < \theta^l_i$.
    \item \textbf{Spike emitted ($s^l_i(t) = 1$):} This occurs if $v^l_i(t-1) + I^l_i(t) \geq \theta^l_i$. Then, $v^l_i(t) = v^l_i(t-1) + I^l_i(t) - \theta^l_i$. Since $v^l_i(t-1) + I^l_i(t) \geq \theta^l_i$, it follows that $v^l_i(t) \geq 0$. Furthermore, given $v^l_i(t-1) < \theta^l_i$ (by induction hypothesis) and $I^l_i(t) < \theta^l_i$ (by premise), we have $v^l_i(t-1) + I^l_i(t) < 2\theta^l_i$. Therefore, $v^l_i(t) = v^l_i(t-1) + I^l_i(t) - \theta^l_i < 2\theta^l_i - \theta^l_i = \theta^l_i$. Thus, $0 \leq v^l_i(t) < \theta^l_i$.
\end{itemize}
In both cases, $0 \leq v^l_i(t) < \theta^l_i$ holds, completing the induction.
\end{proof}

\begin{lemma}
\label{lemma_uniform_equivalence_revised}
Assume a target IF neuron $i$ in layer $l$ receives input spike trains $s_j^{l-1}(t)$ from its incoming input set $\mathcal{S}_i^{l-1}$ with weights $w_{ij}^l$ over $T$ timesteps. 
Let
$k_j^{l-1} = \sum_{t=1}^{T} s_j^{l-1}(t)
$
be the total input spikes from source input $j$.
If the conditions of Lemma~\ref{lemma:bounded_potential_revised} are met, then the SNN output representation of this target neuron can be matched by a QNN neuron whose activation function produces $\lceil \log_2(T+1)\rceil$ distinct levels.
%Assume an IF neuron model receiving input spike trains $s^{l-1}_j(t)$ with weights $w_{ij}^l$ over $T$ timesteps. Let $k_j = \sum_{t=1}^T s^{l-1}_j(t)$ be the total input spikes on channel $j$. If the conditions of Lemma~\ref{lemma:bounded_potential_revised} are met (ensuring $0 \leq v^l_i(t) < \theta^l_i$), then the SNN's output representation ability can be matched by a QNN whose activation function produces $\lceil \log_2(T+1) \rceil$ distinct levels.
\end{lemma}

\begin{proof}
The dynamics of the membrane potential are $v^l_i(t) = v^l_i(t-1) + \sum_j w_{ij}^l s^{l-1}_j(t) - \theta^l_i s^l_i(t)$, with $v_i^l(0)=0$. Summing this relation from $t=1$ to $T$ yields:
\[ \sum_{t=1}^T (v^l_i(t) - v^l_i(t-1)) = \sum_{t=1}^T \sum_j w_{ij}^l s^{l-1}_j(t) - \sum_{t=1}^T \theta^l_i s^l_i(t). \]
The left side is a telescoping sum, $v^l_i(T) - v_i^l(0)$. Let $n^l_i = \sum_{t=1}^T s^l_i(t)$ be the integer count of output spikes, and $k^{l-1}_j = \sum_{t=1}^T s^{l-1}_j(t)$ be the total input spikes from presynaptic neuron $j$. Since $v_i^l(0)=0$, we have:
$ v^l_i(T) = \sum_j w_{ij}^l k^{l-1}_j - \theta^l_i n^l_i. $
Rearranging for $n^l_i$, which is an integer by definition:
\[ \quad n^l_i = \frac{\sum_j w_{ij}^l k^{l-1}_j - v^l_i(T)}{\theta^l_i}. \]
Let $X = \frac{\sum_j w_{ij}^l k^{l-1}_j}{\theta^l_i}$ and $\epsilon = \frac{v^l_i(T)}{\theta^l_i}$. The equation becomes $n^l_i = X - \epsilon$, which implies $X = n^l_i + \epsilon$.
Under the conditions of Lemma~\ref{lemma:bounded_potential_revised}, $0 \leq v^l_i(T) < \theta^l_i$, which implies $0 \leq \epsilon < 1$.
Since $n^l_i$ is an integer and $0 \leq \epsilon < 1$, 
%$\lfloor x \rfloor = m \iff m \le x < m+1$ for integer $m$, directly yields 
$n^l_i = \lfloor X \rfloor$.
Therefore, the number of output spikes is given exactly by:
$n^l_i = \left\lfloor \frac{\sum_j w_{ij}^l k^{l-1}_j}{\theta^l_i} \right\rfloor. $
The average output firing rate of target neuron $i$ in layer $l$ is
$
\phi_i^l = \frac{n_i^l}{T}.
$ An SNN neuron can produce integer spike counts belong to $\{0, 1, \dots, T\}$. This results in $T+1$ distinct output levels.
For an equivalent QNN, let its input activations be $a_j^{l-1} = k^{l-1}_j/T$. The QNN computes a pre-activation $z^l_i = \sum_j w_{ij}^l a_j^{l-1} = (\sum_j w_{ij}^l k^{l-1}_j)/T$. We can build a QNN's activation function $h(z^l_i)$ to map the output of SNN exactly:
\[ h(z^l_i) = \frac{1}{T} \left\lfloor \frac{z^l_i \cdot T}{\theta^l_i} \right\rfloor = \frac{1}{T} \left\lfloor \frac{\sum_j w_{ij}^l k^{l-1}_j}{\theta^l_i} \right\rfloor = \frac{n^l_i}{T}.
%\footnote{The function \(h(z^l_i)\) is the conversion function from real numbers to fixed-point representation during the quantization aware training process. During inference, the weights are quantized to integers and the activation \(z^l_i\) are fix-point values $\in{\frac{N}{T}}$, where $N$ is unsigned integers. So, \(h(z^l_i)\) function is executed as \(z^l_i / \theta^l_i\) and \(\theta^l_i\) is usually chosen as 1 or a power of 2~\cite{10152465,yan2021near}. Then this division can be implemented as a bit shift with negligible energy overhead.} 
\] 
To represent these $T+1$ distinct values, a QNN requires $\lceil \log_2(T+1) \rceil$ bits of precision for its output activations.
\end{proof}

\begin{corollary}
\label{corollary:non_uniform_potential_revised}
%If the approximation made in Lemma~\ref{lemma_uniform_equivalence_revised} regarding the negligible influence of $v^l_i(t)$ does not hold (e.g., if $v^l_i(t)$ is not small relative to $\theta^l_i$ or systematically biased), the direct equivalence is affected. This introduces a form of quantization error or noise, potentially reducing the effective information representation capacity of the SNN or leading to a mismatch with the idealized QNN.

\revise{If the conditions stipulated in Lemma~\ref{lemma:bounded_potential_revised} are violated (e.g., if net input current $I^l_i(t) \ge \theta^l_i$ for some $t$, potentially leading to $v^l_i(t)$ not being constrained within $[0, \theta^l_i)$ in the manner assumed for the exact derivation), the precise relationship $n^l_i = \lfloor (\sum_j w_{ij}^l k^{l-1}_j)/\theta^l_i \rfloor$ may be perturbed. Such deviations can reduce the SNN's encoding precision, ensuring that its useful information representation capacity generally does not exceed that of an QNN with $\lceil \log_2(T+1) \rceil$ bits.}
\end{corollary}

\begin{proof}
\revise{By Lemmas 1 and 2, when the condition in Lemma 1 holds, the spike
count $n_i^l$ exactly matches the corresponding quantized output of
the QNN.

If the residual bound is violated, then the equality above is no longer
guaranteed. In particular, whenever
$
v_i^l(T)\ge \theta_i^l,
$
and thus
$
n_i^l
<
\left\lfloor
\frac{\sum_j w_{ij}^l k_j^{l-1}}{\theta_i^l}
\right\rfloor.
$
%This indicates that the realized SNN output no longer attains the full output alphabet constructed in Lemma 2. 
Hence, its effective
representation capability is lower than that of the ideal
matching case, and therefore cannot exceed that of the corresponding
$\lceil \log_2(T+1)\rceil$-bit QNN neuron.}
\end{proof}

\subsection{Proof for Theorem~2}
\begin{proof}
Let the QNN have $N_{\mathrm{src},i}^{l-1}$ incoming inputs to the target neuron $i$ at layer $l$. The number of incoming inputs with non-zero activations is $(1-\gamma_i^{l-1})N_{\mathrm{src},i}^{l-1}$. Each such non-zero QNN activation $a_j^{l-1}$ is interpreted as an average rate or signal strength.
To represent this input value $a_j^{l-1}$ using rate coding in an SNN over a time window of $T$ discrete timesteps, the corresponding incoming input $j$ of the SNN would receive $k^{l-1}_j = a_j^{l-1} T$ spikes during this window. For incoming inputs  where the QNN activation $a_j^{l-1}=0$, the SNN receives $k^{l-1}_j=0$ spikes.

The total number of input spikes received by the SNN across all $N_{\mathrm{src},i}^{l-1}$ incoming inputs during the window $T$ is the sum of spikes from active incoming inputs:
$ \sum_{j=1}^{N_{\mathrm{src},i}^{l-1}} k^{l-1}_j = \sum_{j \text{ s.t. } a_j^{l-1} \neq 0} (a_j^{l-1} T) = T \sum_{j \text{ s.t. } a_j^{l-1} \neq 0} a_j^{l-1}. $
The average SNN input spike rate $s_{r,i}^{l-1}$ is defined as the total input spikes normalized by the total number of incoming inputs and the number of timesteps:
$
s_{r,i}^{l-1} = \frac{\sum_{j=1}^{N_{\mathrm{src},i}^{l-1}} k^{l-1}_j}{N_{\mathrm{src},i}^{l-1} \cdot T}  = \frac{\sum_{j \text{ s.t. } a_j^{l-1} \neq 0} a_j^{l-1}}{N_{\mathrm{src},i}^{l-1}}.$
Let $N_{\text{active}} = (1-\gamma_i^{l-1})N_{\mathrm{src},i}^{l-1}$ be the number of incoming inputs with non-zero QNN activations. The sum $\sum_{j \text{ s.t. } a_j^{l-1} \neq 0} a_j^{l-1}$ is bounded. Given the assumption that each of the $N_{\text{active}}$ non-zero activations $a_j^{l-1}$ is in the range $[1/T, 1]$:
\begin{itemize}
    \item The minimum sum occurs when all $N_{\text{active}}$ activations take the value $1/T$:
    $ \min \left( \sum_{j \text{ s.t. } a_j^{l-1} \neq 0} a_j^{l-1} \right) = N_{\text{active}} \cdot \frac{1}{T} = (1-\gamma_i^{l-1})N_{\mathrm{src},i}^{l-1} \cdot \frac{1}{T}. $
    \item The maximum sum occurs when all $N_{\text{active}}$ activations take the value $1$:
    $ \max \left( \sum_{j \text{ s.t. } a_j^{l-1} \neq 0} a_j^{l-1} \right) = N_{\text{active}} \cdot 1 = (1-\gamma_i^{l-1})N_{\mathrm{src},i}^{l-1} \cdot 1. $
\end{itemize}
Substituting these bounds into the expression for $s_{r,i}^{l-1}$:
\begin{itemize}
    \item Minimum $s_{r,i}^{l-1} = \frac{(1-\gamma_i^{l-1})N_{\mathrm{src},i}^{l-1}(1/T)}{N_{\mathrm{src},i}^{l-1}} = \frac{1-\gamma_i^{l-1}}{T}$.
    \item Maximum $s_{r,i}^{l-1} = \frac{(1-\gamma_i^{l-1})N_{\mathrm{src},i}^{l-1}(1)}{N_{\mathrm{src},i}^{l-1}} = 1-\gamma_i^{l-1}$.
\end{itemize}
Thus, we establish the bounds $\frac{1-\gamma_i^{l-1}}{T} \leq s_{r,i}^{l-1} \leq 1-\gamma_i^{l-1}$.
\end{proof}

% Based on proof, we analyze three scenarios of spike rates in the following sections given a fixed QNN activation sparsity of $\gamma$:

% \begin{itemize}
%     \item \textbf{Worst-Case for SNN Efficiency (Highest Spike Rate):} This scenario occurs when all $(1-\gamma)N_{\mathrm{src},i}^{l-1}$ active QNN neurons have the maximum activation value, $a_j^{l-1} = 1$ (equivalent to $T/T$). This corresponds to the SNN receiving $T$ spikes for every active input. As derived from Theorem~\ref{theorem_spikerate_revised}, this yields the highest possible average spike rate for the SNN: $s_{r,i}^{l-1, \text{worst}} = 1-\gamma$.

%     \item \textbf{Best-Case for SNN Efficiency (Lowest Spike Rate):} This occurs when all active QNN neurons have the minimum non-zero activation value, $a_j^{l-1} = 1/T$, which corresponds to a single spike event in the SNN for each active input. This results in the lowest possible average spike rate: $s_{r,i}^{l-1, \text{best}} = (1-\gamma)/T$.

%     \item \textbf{Average-Case:} We assume the QNN's non-zero activation values are uniformly distributed within the range $[1/T, 1]$. The average activation value across active neurons is therefore $(1/T + 1)/2$. This leads to an average SNN spike rate of $s_{r,i}^{l-1, \text{avg}} = \frac{(1-\gamma)(1/T + 1)}{2}$.
% \end{itemize}
\subsection{Proof for Proposition~1}
\label{proofC}
% \begin{proof}
% Part (a) follows because Theorem~1 applies independently to each neuron. Applying the same neuron-wise construction to all neurons while preserving the original connectivity yields a QNN twin for the whole network.

% For part (b), Theorem~2 gives
% \[
% \frac{1-\gamma_i^l}{T}\le s_{r,i}^{\,l-1}\le 1-\gamma_i^l
% \]
% for every neuron $(i,l)\in\mathcal{N}$. Multiplying by $N_{\mathrm{src},i}^{l-1}$, summing over all neurons, and normalizing by $\sum_{(i,l)\in\mathcal{N}} N_{\mathrm{src},i}^{l-1}$ yields
% \[
% \frac{1-\gamma}{T}
% \le
% s_r
% \le
% 1-\gamma.
% \]
% \end{proof}

\revise{
\begin{proof}
Part (a) follows by applying Theorem~1 independently to every neuron
while preserving the original connectivity.
For part (b), Theorem~2 holds for every neuron $(i,l)\in\mathcal{N}$.
Multiplying $s_{r,i}^{\,l-1}$
by $N_{\mathrm{src},i}^{l-1}$, summing over all $(i,l)\in\mathcal{N}$,
and normalizing by $\sum_{(i,l)\in\mathcal{N}} N_{\mathrm{src},i}^{l-1}$
gives the conclusion.
\end{proof}
}
\bibliographystyle{IEEEtran}
\bibliography{ref}

\vspace{-12mm}
\begin{IEEEbiography}[{\includegraphics[width=1in,height=1.25in,clip,keepaspectratio]{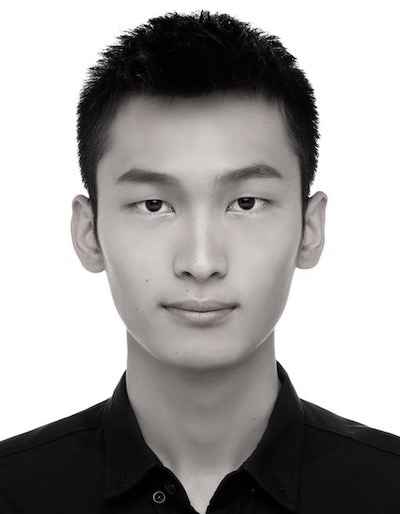}}]{Zhanglu Yan} received the BSc degree in Computer Science from Xi'an Jiaotong University in 2019, the MSc degree in Artificial Intelligence from the National University of Singapore (NUS) in 2020, and the PhD degree in Computer Science from NUS in 2024. He is currently a research fellow at NUS. His research focuses on neuromorphic computing and spiking neural networks.
\end{IEEEbiography}
\vspace{-12mm}
\begin{IEEEbiography}[{\includegraphics[width=1in,height=1.25in,clip,keepaspectratio]{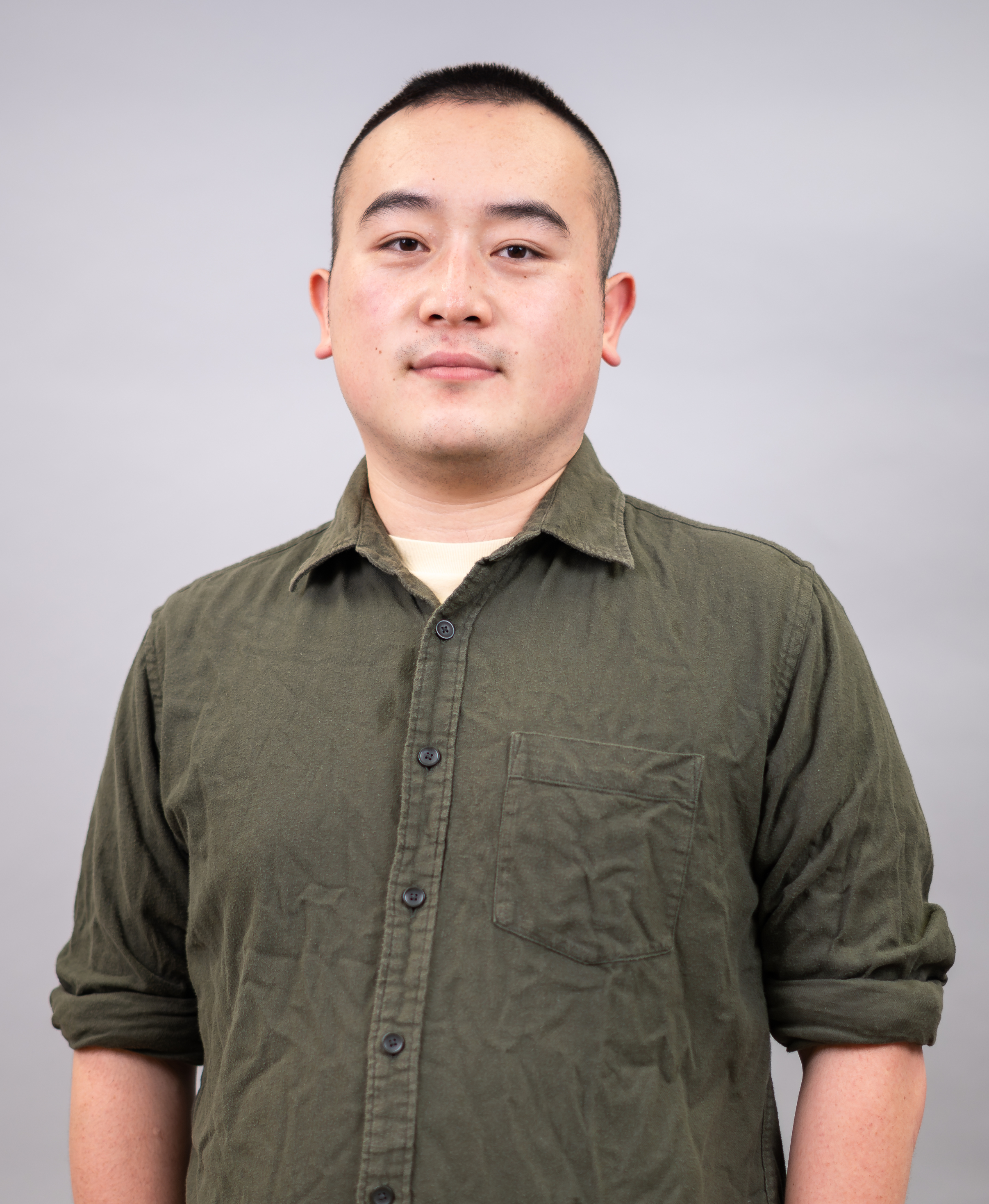}}]{Zhenyu Bai} is a Research Fellow in the Department of Computer Science, National University of Singapore. His research interests include spatial dataflow architectures design and compilation techniques.
\end{IEEEbiography}
\vspace{-12mm}

\begin{IEEEbiography}[{\includegraphics[width=1in,height=1.25in,clip,keepaspectratio]{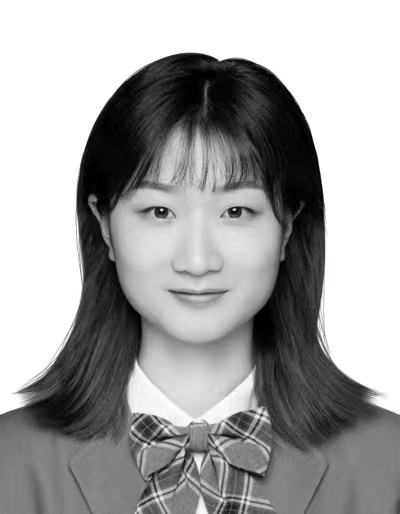}}]{Kaiwen Tang}
Kaiwen Tang received the BSc degree in computer science from Xi'an Jiaotong University in 2022. She is working toward the PhD degree in the School of Computing, National University of Singapore. Her research focuses on spiking neural networks and efficient language models.
\end{IEEEbiography}
\vspace{-12mm}

\begin{IEEEbiography}[{\includegraphics[width=1in,height=1.25in,clip,keepaspectratio]{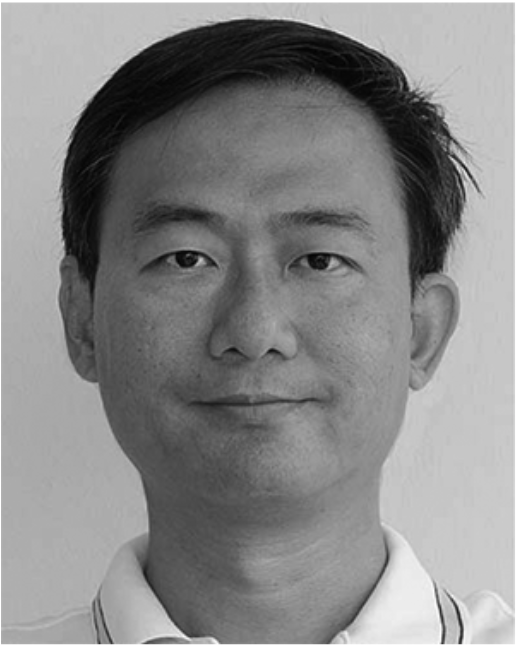}}]{Wong Weng-Fai} received the BSc degree from the National University of Singapore, in 1988, and the DrEngSc degree from the University of Tsukuba, Japan, in 1993. He is currently an associate professor with the Department of Computer Science, National University of Singapore. His research interests include computer architecture, compilers, and high-performance computing. He is a senior member of the IEEE.
\end{IEEEbiography}

% \begin{IEEEbiography}[{\includegraphics[width=1in,height=1.25in,clip,keepaspectratio]{figures/tulika.jpg}}]{Tulika Mitra} is Vice-Provost and Provost’s Chair Professor of Computer Science at the National University of Singapore. Her research focuses on the hardware-software co-design for embedded systems, heterogeneous computing, and Coarse-Grained Reconfigurable Array.
% \end{IEEEbiography}


% % that's all folks

\end{document}